\documentclass{article}

\usepackage{microtype}
\usepackage{graphicx}
\usepackage{subfigure}
\usepackage{booktabs} 

\usepackage{hyperref}
\usepackage{natbib}
\usepackage{adjustbox}
\usepackage{tabularx}
\usepackage{caption}

\usepackage{appendix}

\usepackage[accepted]{icml2024}

\usepackage{amsmath}
\usepackage{amssymb}
\usepackage{mathtools}
\usepackage{amsthm}

\usepackage{tikz}

\usetikzlibrary{decorations}
\usetikzlibrary{decorations.pathmorphing}
\usetikzlibrary{decorations.pathreplacing}
\usetikzlibrary{decorations.shapes}
\usetikzlibrary{decorations.text}
\usetikzlibrary{decorations.markings}
\usetikzlibrary{decorations.fractals}
\usetikzlibrary{decorations.footprints}
\usetikzlibrary{arrows.meta}
\usetikzlibrary{shapes.geometric}
\definecolor{lavande}{RGB}{176,224,230}
\definecolor{lemonchiffon}{RGB}{255,250,205}
\definecolor{burgundy}{rgb}{0.5, 0.0, 0.13}

\usepackage[capitalize,noabbrev]{cleveref}

\theoremstyle{plain}

\theoremstyle{definition}

\theoremstyle{remark}

\usepackage[textsize=tiny]{todonotes}

\icmltitlerunning{Beyond Individual Input for Deep Anomaly Detection on Tabular Data}

\begin{document}

\twocolumn[
\icmltitle{Beyond Individual Input for Deep Anomaly Detection on Tabular Data}




\begin{icmlauthorlist}
\icmlauthor{Hugo Thimonier}{lisn}
\icmlauthor{Fabrice Popineau}{lisn}
\icmlauthor{Arpad Rimmel}{lisn}
\icmlauthor{Bich-Li\^en Do\^an}{lisn}
\end{icmlauthorlist}

\icmlaffiliation{lisn}{Université Paris-Saclay, CNRS, CentraleSup\'elec, Laboratoire Interdisciplinaire des Sciences du Num\'erique, 91190, Gif-sur-Yvette, France.}

\icmlcorrespondingauthor{Hugo Thimonier}{name.surname@lisn.fr}

\icmlkeywords{Machine Learning, Anomaly Detection, Tabular Data}

\vskip 0.3in
]



\printAffiliationsAndNotice{\icmlEqualContribution} 

\begin{abstract}
Anomaly detection is vital in many domains, such as finance, healthcare, and cybersecurity. In this paper, we propose a novel deep anomaly detection method for tabular data that leverages Non-Parametric Transformers (NPTs), a model initially proposed for supervised tasks, to capture both feature-feature and sample-sample dependencies. In a reconstruction-based framework, we train an NPT to reconstruct masked features of normal samples. In a non-parametric fashion, we leverage the whole training set during inference and use the model's ability to reconstruct the masked features to generate an anomaly score. To the best of our knowledge, this is the first work to successfully combine feature-feature and sample-sample dependencies for anomaly detection on tabular datasets. Through extensive experiments on 31 benchmark tabular datasets, we demonstrate that our method achieves state-of-the-art performance, outperforming existing methods by 2.4\% and 1.2\% in terms of F1-score and AUROC, respectively. Our ablation study further proves that modeling both types of dependencies is crucial for anomaly detection on tabular data.
\end{abstract}

\section{Introduction}
    \label{introduction}

        Anomaly detection is a critical task that aims to identify samples that deviate from a pre-defined notion of normality within a dataset. Traditional approaches to anomaly detection characterize the \textit{normal}\footnote{The term \textit{normal} here relates to the concept of normality in opposition to \textit{abnormal}.} distribution almost exclusively using samples considered as \textit{normal}, and flag data points as anomalies based on their deviation from this distribution. 
        Anomaly detection (AD) is especially useful for applications involving imbalanced datasets, where standard supervised methods may fail to achieve satisfactory performance \cite{Yanminsun2011}.
        Those applications include fraud detection \cite{hilalfraud}, intrusion detection in cybersecurity \cite{Malaiya2018} or astronomy \cite{astrology}.
    
        Anomaly detection encompasses both unsupervised and semi-supervised methods. In most real-world scenarios, labeled datasets that differentiate normal samples from anomalies are unavailable or costly to obtain. To address this, efficient anomaly detection methods must be robust to dataset contamination, where the training set is predominantly composed of normal samples but also includes anomalies. 
        However, when labeled data is available, one can consider a semi-supervised approach to create a training set consisting solely of \textit{normal} samples, thereby indirectly incorporating label information into the anomaly detection model.

        Many general AD methods tend to work well on tasks that involve unstructured data (\textit{e.g.}, natural language processing or computer vision) such as \cite{ocsvm, svdd, isolationforest, deep-svdd, Kim2020RaPP, liznerski2021explainable}. However, recent work \cite{goad, neutralad, shenkar2022anomaly} has revealed that the best-performing methods for tabular data involve models tailored to consider the particular structure of this data type. AD methods for structured data typically identify anomalies by using either \textit{feature-feature} or \textit{sample-sample} dependencies. For instance, in \cite{shenkar2022anomaly}, the authors assume a class-dependent relationship between a subset of variables in a sample's feature vector and the rest of its variables.
        The authors thus propose a contrastive learning framework to detect anomalies based on this assumption. 
        Another recent method \cite{thimonier2022} identifies anomalies in tabular datasets by focusing on sample-sample dependencies by measuring the influence of training \textit{normal} samples on the validation samples.
        Both approaches have demonstrated competitive results for anomaly detection in tabular datasets.
    
        Recent work on supervised deep learning methods for tabular data \cite{Shavitt2018, tabnet2019, gorishniy2023tabr, saint2021, kossen2021self} has also highlighted the importance of considering the particular structure of tabular data.
        In particular, in \cite{kossen2021self, saint2021, gorishniy2023tabr}, the authors emphasize the significance of considering both feature-feature and sample-sample dependencies for supervised regression and classification problems on tabular data. 
        Based on the latter observation, we hypothesize that feature-feature relations and sample-sample dependencies are class-dependent, and \textbf{both dependencies should be used conjointly to identify anomalies}. In particular, since interactions between samples are learned exclusively using \textit{normal} samples in the anomaly detection setup, they should be especially discriminative in identifying anomalies during inference.

        To test this hypothesis, we employ Non-Parametric Transformers (NPT) \cite{kossen2021self}, first proposed for supervised tasks on tabular datasets.
        We show that NPTs are particularly relevant for flagging anomalies, in line with recent work \cite{han2022adbench} demonstrating the effectiveness of new deep learning architectures for anomaly detection on tabular data. We experiment on an extensive benchmark of tabular datasets to demonstrate the capacity of our approach to detect anomalies and compare our performances to existing AD methods. We obtain state-of-the-art results when it comes to detection accuracy. We also test the robustness of our approach to dataset contamination and give evidence that it can serve for unsupervised anomaly detection when the training set contamination is not too severe.
        Finally, our ablation study, conducted with reconstruction-based approaches similar to our proposed method but utilizing K-nearest neighbors (KNN) imputation and a vanilla transformer \cite{attentionisallyouneed}, provides evidence that considering both types of dependencies can be crucial to detect anomalies on specific datasets accurately.

        The present work offers the following contributions:
        \begin{itemize}
        \itemsep0em 
            \item[•] We propose the first AD method for tabular data relying on masked feature reconstruction.
            \item[•] We put forward the first deep AD method to successfully combine feature-feature and sample-sample dependencies.
            \item[•] Our method shows state-of-the-art anomaly detection capacity on an extensive benchmark of 31 tabular datasets.
            \item[•] We provide strong evidence of the crucial role of considering both dependencies for anomaly detection on tabular data.
        \end{itemize}

    \section{Related works}
    \label{related-works}
       
        Anomaly detection approaches can be categorized into four main types: density estimation, one-class classification, reconstruction-based, and self-supervised.
        \paragraph{Density Estimation} The most straightforward approach to detecting samples that do not belong to a distribution is to estimate the distribution directly and to measure the likelihood of a sample under the estimated distribution. Several approaches found in the literature have considered using non-parametric density estimation methods to estimate the density of the \textit{normal} distribution, such as KDE \cite{Parzen}, GMM \cite{gmm}, or Copula as in COPOD \cite{Li2020}. Other approaches also focused on local density estimation to detect outliers, such as Local Outlier Factor (LOF) \cite{lof}. In inference, one flags the samples that lie in low-probability regions under the estimated distribution as anomalies.
        
        \paragraph{Reconstruction Based Methods} Other methods have consisted in learning to reconstruct samples that belong to the \textit{normal} distribution. In this framework, the models' incapacity to reconstruct a sample correctly serves as a proxy to measure anomaly. A high reconstruction error would indicate that a sample does not belong to the estimated \textit{normal} distribution. Those approaches can involve PCA \cite{pca} or neural networks such as diverse types of autoencoders \cite{Principi, chen2018unsupervised, Kim2020RaPP}, or GANs \cite{anogan}.

        \paragraph{One-Class Classification} The term \textit{one-class classification} was coined in \cite{occ} and describes identifying anomalies without directly estimating the \textit{normal} density. One-class classification (OCC) involves discriminative models that directly estimate a decision boundary. For instance, in kernel-based approaches \cite{ocsvm, svdd}, authors propose to characterize the support of the \textit{normal} samples in a Hilbert space and to flag as anomalies the samples that would lie outside of the estimated support. Similarly, recent work has extended their approach by replacing kernels with deep neural networks \cite{deep-svdd}.
        In \cite{goyalDROCCDeepRobust2020}, authors proposed DROCC that involves generating, in the course of training,  synthetic anomalous samples in order to learn a classifier on top of the one-class representation. Other OCC approaches have relied on tree-based models such as isolation forest (IForest) \citep{isolationforest}, extended isolation forest \citep{ief}, RRCF \citep{Guha2016} and PIDForest \citep{Gopalan2019PIDForestAD}.
        
        \paragraph{Self-Supervised Approaches} Recent methods have also considered self-supervision as a means to identify anomalies. In GOAD \cite{goad}, authors apply several affine transformations to each sample and train a classifier to identify from the transformed samples which transformation was applied. The classifier only learns to discriminate between transformations using \textit{normal} transformed samples: assuming this problem is class-dependent, the classifier should fail to identify transformation applied to anomalies. In NeuTraL-AD \cite{neutralad}, authors propose a contrastive framework in which samples are transformed using neural mappings and are embedded in a latent semantic space using an encoder. The objective is to learn transformations so that transformed samples still share similarities with their untransformed counterpart while different transformations are easily distinguishable. The contrastive loss then serves as the anomaly score in inference. 
        Similarly, \cite{shenkar2022anomaly} also propose a contrastive framework in which they identify samples as anomalies based on their inter-feature relations. Other self-supervised approaches, such as \cite{sohn2021learning, reiss2021mean}, have focused on representation learning to foster the performance of one-class classification models.
        
        \paragraph{Attention Mechanisms} First introduced in \cite{attentionisallyouneed}, the concept of attention has become ubiquitous in the machine learning literature. Scholars have successfully applied transformers on a broad range of tasks, including computer vision, \textit{e.g.} image generation with the Image Transformer \cite{imagetransformer} or image classification with the Vision Transformer (ViT) \cite{dosovitskiy2021an}, natural language processing \textit{e.g.} Masked Language Models (MLM) such as BERT \cite{DevlinCLT19}, and classification tasks on structured datasets \cite{saint2021, kossen2021self}.
        
        \paragraph{Deep Learning for Tabular Data} 
        Despite the effectiveness of deep learning models for numerous tasks involving unstructured data, non-deep models remain the prevalent choice for machine learning tasks such as classification and regression on tabular data \cite{grinsztajn2022why, shwartz-ziv2021tabular}. However, in recent years, scholars have shown that one could successfully resort to deep learning methods for various tasks on tabular datasets. For instance, in \cite{Shavitt2018, jeffares2023tangos}, authors discuss how regularization is crucial in training a deep learning model tailored for tabular data. Hence, they propose a new regularization loss to accommodate the variability between features.
        Similarly, \cite{kadra2021well} shows that correctly selecting a combination of regularization techniques can suffice for a Multi-Layer Perceptron (MLP) to compete with GBDT. Finally, \cite{saint2021, kossen2021self} propose deep learning models based on attention mechanisms that rely on feature-feature, feature-label, sample-sample, and sample-label attention. Both models achieve competitive results on several baseline datasets and emphasize sample-sample interaction's role in classifying samples correctly.

    \section{Method}
    \label{sec:method}
    
        In this section, we discuss the learning objective used to optimize the parameters of our model, then we briefly present the mechanisms involved in Non-Parametric Transformers \cite{kossen2021self}, and finally, we present NPT-AD, our method to derive an anomaly score.
    
        \subsection{Learning Objective}
        \label{subsec:learning_objective}
    
            Reconstruction-based approaches for anomaly detection involve training a model to accurately reconstruct \textit{normal} samples while failing to reconstruct anomaly samples. Such methods effectively identify anomalies by exploiting differences in the underlying data distributions between \textit{normal} and anomalous samples.
            Let $\mathcal{D}_{train} = \{\mathbf{x}_i \in \mathbb{R}^{d}\}_{i=1}^n$ represent the training set composed of $n$ \textit{normal} samples with $d$ features. Standard reconstruction-based approaches consider learning a mapping $\phi_\theta:\mathbb{R}^d \to \mathbb{R}^d$ to minimize a reconstruction loss. The parameters $\theta \in \Theta$ are optimized to reconstruct each sample $\mathbf{x} \in \mathbb{R}^d$ in the training set with minimal error. Formally, the overall objective can be expressed as
            \begin{equation}
                \label{eq:recon}
                \min_{\theta \in \Theta} \sum_{\mathbf{x} \in \mathcal{D}_{train}} d(\mathbf{x}, \phi_\theta(\mathbf{x})),
            \end{equation}
            where $d(\mathbf{x},\phi_\theta(\mathbf{x}))$ measures how well the model reconstructs sample $\mathbf{x}$. The latter is often set to be a distance measure such as the Euclidean distance. 
            
            The AD method proposed in \cite{shenkar2022anomaly} employs a masking strategy that maximizes the mutual information between each sample and its masked-out part by minimizing a contrastive loss. Recently, \cite{Kong2020A} demonstrated how stochastic masking \cite{DevlinCLT19} also maximizes mutual information, thereby establishing a link between the method of \cite{shenkar2022anomaly} and stochastic masking. 
            In stochastic masking, each entry in a sample vector $\mathbf{x} \in \mathbb{R}^d$ is masked with probability $p_{mask}$, and the objective task is to predict the masked-out features from the unmasked features. 
             Formally, let $\mathbf{m}\in \mathbb{R}^{d}$ be a binary vector taking value $1$ when the corresponding entry in $\mathbf{x}$ is masked, $0$ otherwise. Let $\mathbf{x}^m, \mathbf{x}^o \in \mathbb{R}^d$ represent respectively the masked and unmasked entries of sample $x$ defined as
                \begin{equation}
                    \begin{array}{rl}
                         \mathbf{x}^m & = \mathbf{m} \odot \mathbf{x}  \\
                         \mathbf{x}^o  & = (\mathbf{1}_d-\mathbf{m}) \odot \mathbf{x},
                    \end{array}
                \end{equation}
                where $\mathbf{1}_d$ is the $d$-dimensional unit vector.
                
                In this framework, the objective in eq. \ref{eq:recon} is modified to
            \begin{equation}
                \label{eq:masked_recon}
                \min_{\theta \in \Theta} \sum_{\mathbf{x} \in \mathcal{D}_{train}} d(\mathbf{x}^m, \phi_\theta(\mathbf{x}^o)),
            \end{equation}
            where $\phi_\theta(\mathbf{x}^o)$ denotes the reconstructed masked features of sample $\mathbf{x}$ by the model.
        
        
            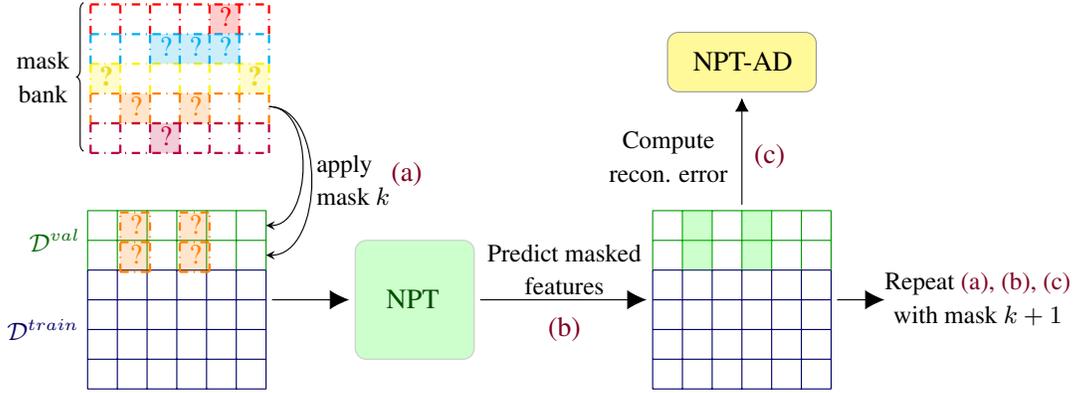
\begin{figure*}[t!]
            \begin{center}
            \resizebox{0.85\textwidth}{!}{
                \begin{tikzpicture}[scale=0.75]
                    \draw[step=0.5, color=blue!40!black] (0,0) grid (3, 2);
                    \draw[step=0.5, color=green!60!black] (0,2) grid (3, 3);
                    \draw[orange, dashdotted, step=0.5cm, thick, xshift=0.05cm, yshift=-0.03cm] (0.5-0.001,2-0.001) grid (1,3);
                    \draw[orange, dashdotted, step=0.5cm, thick, xshift=0.05cm, yshift=-0.03cm] (1.5-0.001,2-0.001) grid (2,3);
                    \node[orange ,anchor = center, xshift=0.05cm, yshift=0.01cm] () at (.75,2.25){?};
                    \node[orange ,anchor = center, xshift=0.05cm, yshift=0.01cm] () at (.75,2.75){?};
                    \node[orange ,anchor = center, xshift=0.05cm, yshift=0.01cm] () at (1.75,2.25){?};
                    \node[orange ,anchor = center, xshift=0.05cm, yshift=0.01cm] () at (1.75,2.75){?};
                    \path [anchor = center, fill=orange, opacity=0.2, xshift=0.05cm, yshift=-0.03cm] (0.5,2) rectangle (1,3);
                    \path [anchor = center, fill=orange, opacity=0.2, xshift=0.05cm, yshift=-0.03cm] (1.5,2) rectangle (2,3);
                    \path (-0.55, 2.5) node[color=green!40!black] {\small{$\mathcal{D}^{val}$}};
                    \path (-0.75, 1) node[color=blue!40!black] {\small{$\mathcal{D}^{train}$}};
                    
                    \draw[orange, dashdotted, step=0.5cm, thick, xshift=0.05cm, yshift=-0.03cm] (0-0.001,4.5-0.001) grid (3,5);
                    \path [anchor = center, fill=orange, opacity=0.2, xshift=0.05cm, yshift=-0.03cm] (0.5,4.5) rectangle (1,5);
                    \path [anchor = center, fill=orange, opacity=0.2, xshift=0.05cm, yshift=-0.03cm] (1.5,4.5) rectangle (2,5);
                    \node[orange ,anchor = center, xshift=0.05cm, yshift=0.01cm] () at (.75,4.75){?};
                    \node[orange ,anchor = center, xshift=0.05cm, yshift=0.01cm] () at (1.75,4.75){?};
                    
                    \draw[purple, dashdotted, step=0.5cm, thick, xshift=0.05cm, yshift=-0.03cm] (0-0.001,4-0.001) grid (3,4.5);
                    \path [anchor = center, fill=purple, opacity=0.2, xshift=0.05cm, yshift=-0.03cm] (1,4) rectangle (1.5,4.5);
                    \node[purple ,anchor = center, xshift=0.05cm, yshift=0.01cm] () at (1.25,4.25){?};
                    
                    \draw[yellow, dashdotted, step=0.5cm, thick, xshift=0.05cm, yshift=-0.03cm] (0,5+0.001) grid (3,5.5+0.001);
                    \path [anchor = center, fill=yellow, opacity=0.2, xshift=0.05cm, yshift=-0.03cm] (0,5) rectangle (0.5,5.5);
                    \path [anchor = center, fill=yellow, opacity=0.2, xshift=0.05cm, yshift=-0.03cm] (2.5,5) rectangle (3,5.5);
                    \node[yellow!90!black, anchor = center, xshift=0.05cm, yshift=0.01cm] () at (.25,5.25){\textbf{?}};
                    \node[yellow!90!black, anchor = center, xshift=0.05cm, yshift=0.01cm] () at (2.75,5.25){\textbf{?}};
                    
                    \draw[cyan, dashdotted, step=0.5cm, thick, xshift=0.05cm, yshift=-0.03cm] (0,5.5) grid (3,6);
                    \path[cyan, anchor = center, fill=cyan, opacity=0.2, xshift=0.05cm, yshift=-0.03cm] (1,5.5) rectangle (2.5,6);
                    \node[cyan, anchor = center, xshift=0.05cm, yshift=0.01cm] () at (1.25,5.75){?};
                    \node[cyan, anchor = center, xshift=0.05cm, yshift=0.01cm] () at (1.75,5.75){?};
                    \node[cyan, anchor = center, xshift=0.05cm, yshift=0.01cm] () at (2.25,5.75){?};
                    
                    \draw[red, dashdotted, step=0.5cm, thick, xshift=0.05cm, yshift=-0.03cm] (0,6) grid (3,6.5);
                    \path[red, anchor = center, fill=red, opacity=0.2, xshift=0.05cm, yshift=-0.03cm] (2,6) rectangle (2.5,6.5);
                    \node[red, anchor = center, xshift=0.05cm, yshift=0.01cm] () at (2.25,6.25){?};
                    \draw[color=black ,decorate,decoration={brace,raise=0.05cm}] (0,4) -- (0,6.5);
                
                    \path[align=center] (-0.75, 5.25) node[color=black] {\footnotesize{mask} \\ \footnotesize{bank}};
                    \draw[->,>=stealth, color=black](3.05, 4.75) to [out=0,in=0] (3, 2.75);
                    \draw[->,>=stealth, color=black](3.05, 4.75) to [out=0,in=0] (3, 2.25);
                    \node[align=left, color=black, anchor = center, xshift=0.05cm, yshift=0.01cm] () at (4.4,3.5){\footnotesize{apply} \\ \footnotesize{mask $k$}};
                    \node[align=left, color=burgundy, anchor = center, xshift=0.05cm, yshift=0.01cm] () at (5.3,3.6){(a)};
                
                    \draw[rounded corners, fill=green, opacity=0.2] (4.5, 0.5) rectangle (6.5,2.5) {};
                    \node[align=center, anchor = center, xshift=0.05cm, yshift=0.01cm, color=green!30!black] () at (5.4,1.5){NPT};
        
                    \draw [arrows = {-Latex[width=7pt, length=7pt]}] (3.1,1.5) -- (4.4,1.5);
                    \draw [arrows = {-Latex[width=7pt, length=7pt]}] (6.6,1.5) -- (9.4,1.5);
                    \node[align=center, anchor = center, xshift=0.05cm, yshift=0.01cm] () at (7.95,2){\footnotesize{Predict masked}\\ \footnotesize{features}};
                    \node[align=left, color=burgundy, anchor = center, xshift=0.05cm, yshift=0.01cm] () at (7.95,1){(b)};
        
                    \draw[step=0.5, color=blue!40!black] (9.5-0.001,0) grid (12.5,2);
                    \draw[step=0.5, color=green!60!black] (9.5-0.001,2) grid (12.5,3);
                    
                    \path [color=black, anchor = center, fill=green, opacity=0.2] (10,2) rectangle (10.5,3);
                    \path [anchor = center, fill=green, opacity=0.2] (11,2) rectangle (11.5,3);
                    \draw [arrows = {-Latex[width=5pt, length=5pt]}] (11,3.1) -- (11,4.9);
        
                    \draw[rounded corners, fill=yellow, opacity=0.4] (9.75, 5) rectangle (12.25,6) {};
                    \node[align=center, anchor = center, xshift=0.05cm, yshift=0.01cm] () at (10.95,5.5){NPT-AD};
                    \node[align=center, anchor = center, xshift=0.05cm, yshift=0.01cm] () at (9.7,3.9) {\footnotesize{Compute} \\ \footnotesize{recon. error}};
                    \node[align=left, color=burgundy, anchor = center, xshift=0.05cm, yshift=0.01cm] () at (11.4,3.9){(c)};
        
                    \draw [arrows = {-Latex[width=7pt, length=7pt]}] (12.6,1.5) -- (13.4, 1.5);
                    \node[align=center, anchor = center, xshift=0.05cm, yshift=0.01cm] () at (14.9,1.5) {\footnotesize{Repeat \textcolor{burgundy}{(a), (b), (c)}} \\ \footnotesize{with mask $k+1$}}
                    ;
                \end{tikzpicture}
                }
                \end{center}
                \caption{NPT-AD inference pipeline. In step \textcolor{burgundy}{(a)}, mask $j$ is applied to each validation sample. We construct a matrix $\mathbf{X}$ composed of the masked validation samples and the whole \textit{unmasked} training set. In step \textcolor{burgundy}{(b)}, we feed $\mathbf{X}$ to the Non-Parametric Transformer (NPT), which tries to reconstruct the masked features for each validation sample. On top of the learned feature-feature interactions,  NPT will use the unmasked training samples to reconstruct the mask features. In step \textcolor{burgundy}{(c)}, we compute the reconstruction error that we later aggregate in the NPT-AD score.}
                \label{fig:inference}
                \end{figure*}

            Our proposed approach leverages the entire dataset in a non-parametric manner to reconstruct masked features. This method considers feature-feature interactions and also captures relationships between samples to optimize the reconstruction objective. Let $\mathbf{X} \in \mathbb{R}^{n \times d}$ denote the dataset matrix, consisting of $n$ training samples with $d$ features. We introduce the matrix equivalents of $m$, $\mathbf{x}^m$, and $\mathbf{x}^o$, denoted as $\mathbf{M}$, $\mathbf{X}^M$, and $\mathbf{X}^O$, respectively, all in $\mathbb{R}^{n \times d}$.  
            The reconstruction objective described in eq. \ref{eq:masked_recon} can then be reformulated as
            \begin{equation}
            \label{eq:masked_recon_nptad}
            \min_{\theta \in \Theta} \sum_{\mathbf{x} \in \mathcal{D}_{train}} d \left(\mathbf{x}^m, \phi_\theta\left( \mathbf{x}^o\mid \mathbf{X}^O \right) \right).
            \end{equation}    
    
        \subsection{Non-parametric transformer (NPT)}
        \label{subsec:npt}
    
            We resort to Non-Parametric Transformer (NPT) \cite{kossen2021self} as the core model for our approach, denoted as $\phi_\theta$ in section \ref{subsec:learning_objective}. NPT involves both attention between features and attention between samples, thus allowing the ability to capture feature-feature and sample-sample dependencies. More precisely, two mechanisms involved in NPTs allow anomalies to be identified: Attention Between Datapoints (ABD) and Attention Between Attributes (ABA). 
            Both attention mechanisms rely on multi-head self-attention (MHSA), which was first introduced in the natural-language processing literature \cite{Bahdanau2014, DevlinCLT19, attentionisallyouneed}. We discuss MHSA more thoroughly in appendix \ref{appendix:mhsa} and only detail in this section the two mechanisms put forward in \cite{kossen2021self}.
        
            As an input, NPT receives both the dataset and a masking matrix $(\mathbf{X}, \mathbf{M}) \in \mathbb{R}^{n\times d} \times \mathbb{R}^{n\times d}$. Before feeding the input to the NPT, we pass each of the $n$ data samples through a linear embedding layer to obtain an $e$-dimensional embedding for each feature. Thus, as an input, NPT receives a representation $\mathbf{H}^0 \in \mathbb{R}^{n\times d \times e}$. A sequence of MHSA layers is applied to the input, alternating between ABA and ABD. The model then outputs a prediction for masked features while keeping unmasked features unchanged $\widehat{\mathbf{X}} \in \mathbb{R}^{n\times d}$. 
        
            \paragraph{Attention Between Datapoints (ABD)} It is the key feature that differentiates NPT from standard transformer models. This mechanism captures pairwise relations between data samples. Consider as an input to the ABD layer the previous layer representation $\mathbf{H}^{(\ell)} \in \mathbb{R}^{n\times d \times e}$ flattened to $\mathbb{R}^{n\times h}$ where $h=d\cdot e$. Then, NPT applies MHSA, as seen in equation \ref{eq:mhsa} in appendix \ref{appendix:mhsa}, between the data samples flattened representations $\{\mathbf{H}_i^{(\ell)} \in \mathbb{R}^{1\times h} | i\in 1,\dots,n\}$. 
            \begin{equation}
                \mbox{ABD}(\mathbf{H}^{(\ell)}) = \mbox{MHSA}(\mathbf{H}^{(\ell)}) = \mathbf{H}^{(\ell+1)} \in \mathbb{R}^{n\times h}
            \end{equation}
            After applying ABD, the data representation is reshaped to its original dimension in $\mathbb{R}^{n\times d \times e}$.
        
            \paragraph{Attention Between Attributes (ABA)} As already discussed, NPT alternates between ABD and ABA layers. ABA layers should help learn per data sample representation for the inter-sample representations. In contrast with ABD, ABA consists in applying MHSA independently to each row in $\mathbf{H}^{(\ell)}$, \textit{i.e.} to each data sample's intermediate representation $\mathbf{H}_i^{(\ell)} \in \mathbb{R}^{d\times e}, i\in \{1,\dots,n\}$.
            \begin{equation}
                \mbox{ABA}(\mathbf{H}^{(\ell)}) = \underset{\tiny \mbox{axis=n}}{\mbox{stack}}\left(\mbox{MHSA}(\mathbf{H}_1^{(\ell)}), \dots, \mbox{MHSA}(\mathbf{H}_n^{(\ell)}) \right)
            \end{equation}

        \subsection{Anomaly score}
        \label{subsec:anomalyscore}

            We directly derive the anomaly score from the loss optimized during training. The loss corresponds to the squared difference between the reconstructed feature and its actual value for numerical features. Meanwhile, for categorical features, we use the cross-entropy loss function. The anomaly score relies on our model's capacity to reconstruct masked features correctly and assumes that the model should better reconstruct \textit{normal} samples. 
            Two reasons support this assumption. First, relations between features are class-dependent
            , having observed only \textit{normal} samples in the training phase, the model should be unable to fetch the learned feature-feature interactions to reconstruct anomalies properly. Second,  sample-sample interactions seen by the model only correspond to interactions between \textit{normal} samples, making it difficult to successfully exploit interactions between \textit{normal} samples and anomalies.
            
            As detailed in Figure \ref{fig:inference}, we consider a mask bank composed of $m$ $d$-dimensional deterministic mask vectors that designate which of the $d$ features of \textit{each} validation sample will be hidden. 
            We set the maximum number of features to be masked simultaneously $r$, and construct $m=\sum_{k=1}^r \binom{d}{k}$ masks.
            Each mask is applied to each validation sample $\mathbf{z} \in \mathcal{D}^{val}$ to obtain $m$ different masked samples $\{\mathbf{z}^{(1)},\dots,\mathbf{z}^{(m)}\}$ of the original sample $\mathbf{z}$.
            We use the whole unmasked training set\footnote{For large datasets, we resort to a random subsample of the training set for computational reasons.} $\mathcal{D}^{train}$ to predict the masked features of each sample for each of the $m$ masked vectors and construct the anomaly score for a validation sample $\mathbf{z}$ as
            \begin{equation}
                \label{eq:npt-ad}
                \mbox{NPT-AD}(\mathbf{z}; \mathcal{D}^{train}) = \frac{1}{m} \sum_{k=1}^m \mathcal{L}(\mathbf{z}^{(k)};\mathcal{D}^{train}), 
            \end{equation}
            where $\mathcal{L}(\mathbf{z}^{(k)};\mathcal{D}^{train})$ designates the loss for the sample $\mathbf{z}$ with mask $k$. We also considered other forms of aggregation, such as the maximum loss over all masks.
    
    \section{Experiments}
    \label{sec:experiments}

        \paragraph{Datasets} We experiment on an extensive benchmark of tabular datasets following previous work \cite{shenkar2022anomaly}. The benchmark is comprised of two datasets widely used in the anomaly detection literature, namely Arrhythmia and Thyroid, a second group of datasets, the "Multi-dimensional point datasets", obtained from the Outlier Detection DataSets (ODDS)\footnote{\url{http://odds.cs.stonybrook.edu/}} containing 28 datasets. We omit the datasets Heart and Yeast following the literature and also omit the KDD dataset since it presents a certain number of limitations \cite{kddnogood}. Instead, we include three real-world datasets from \cite{han2022adbench} that display relatively similar characteristics to KDD in terms of dimensions: fraud, campaign, and backdoor. See appendix \ref{appendix:exp_settings} for more detail on the datasets' characteristics. 

        \paragraph{Experimental settings} Per the literature \cite{zong2018deep, goad}, we construct the training set with a random subsample of the \textit{normal} samples representing 50\% of the \textit{normal} samples, we concatenate the 50\% remaining with the entire set of anomalies to constitute the validation set. 
        Following previous work, \cite{goad}, the decision threshold for the NPT-AD score is chosen such that the number of predicted anomalies is equal to the number of existing anomalies. 
        We report the results in tables \ref{tab:odds_f1-1}, \ref{tab:odds_f1-2} \ref{tab:odds_auc} and \ref{tab:odds_auc_2} in appendix \ref{appendix:add_rez}. Most metrics are obtained from \cite{shenkar2022anomaly}, apart from NeuTraL-AD \cite{neutralad}, which we trained using their official code made available online, the transformer model, and the experiments on the fraud, campaign, and backdoor datasets. Per the literature, we evaluate the different methods using the F1-score ($\uparrow$) and AUROC ($\uparrow$) metrics. We compare our method to recent deep methods, namely GOAD \cite{goad}, DROCC \cite{goyalDROCCDeepRobust2020}, NeuTraL-AD \cite{neutralad}, the contrastive approach proposed in \cite{shenkar2022anomaly} and a transformer model \cite{attentionisallyouneed} trained in a similar way to NPT-AD. We also compare to classical non-deep methods such as Isolation Forest \cite{isolationforest}, KNN \cite{ramaswamy2000efficient}, RRCF \cite{Guha2016}, COPOD \cite{Li2020} and PIDForest \cite{Gopalan2019PIDForestAD}. We refer the reader to \cite{shenkar2022anomaly} for implementation details of non-deep models.
        \begin{figure*}[t]
            \centering
            
              \subfigure[F1-score ($\uparrow$)]{
                \includegraphics[width=0.48\linewidth]{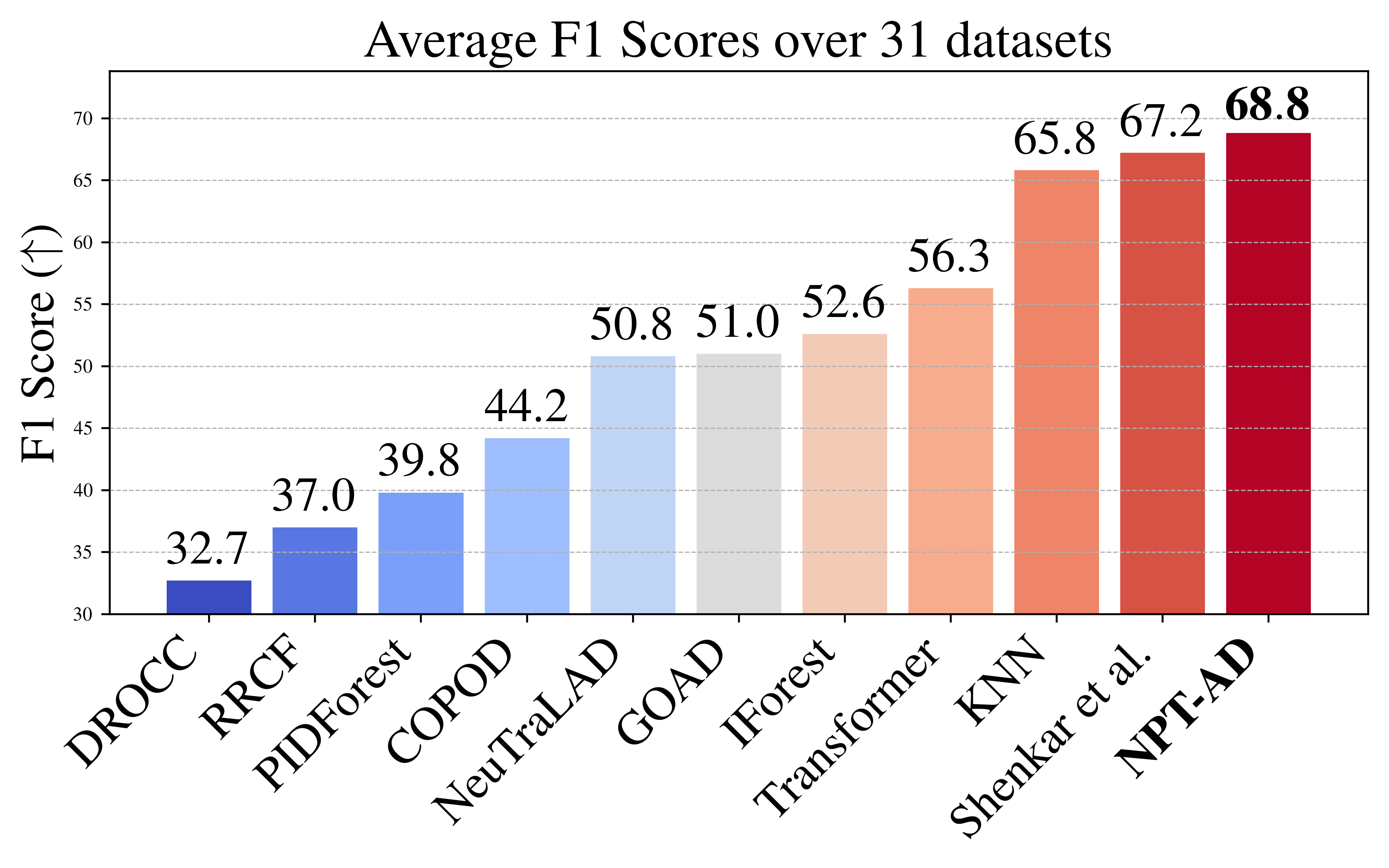}
                \label{fig:sub_avg_F1}
              }
              \hfill
              \subfigure[Rank ($\downarrow$)]{
                \includegraphics[width=0.48\linewidth]{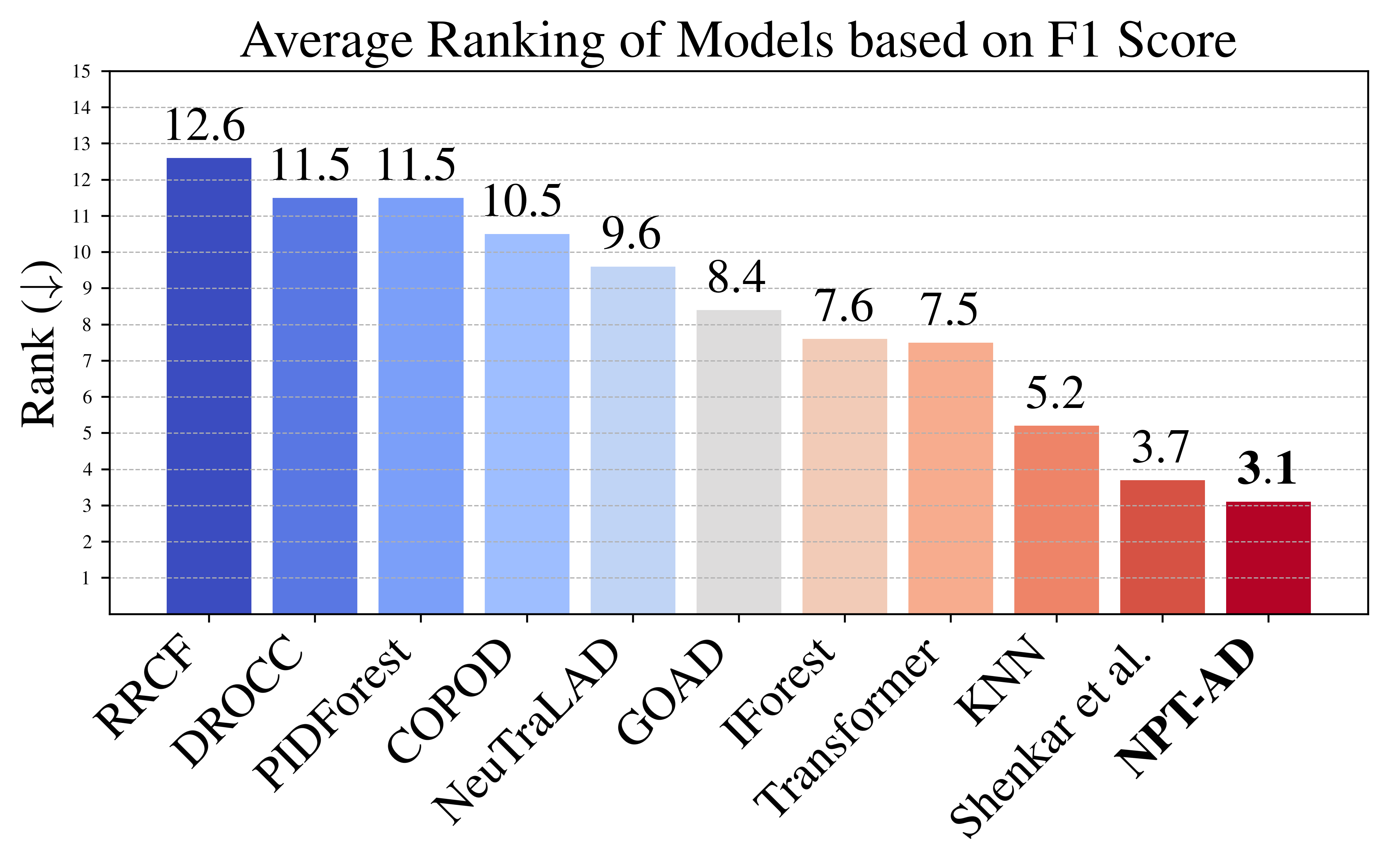}
                \label{fig:sub_avg_rk}
            }
              \caption{For each of the $31$ datasets on which models were evaluated, we compute the average F1-score over $20$ runs for $20$ different seeds. We report on figure \ref{fig:sub_avg_F1} the average F1-score over all datasets for each tested model. We report on figure \ref{fig:sub_avg_rk} the average rank over the 31 datasets. For both figures, the model displayed on the far left is the worst-performing model for the chosen metric, and on the far right is the best-performing model. We also highlight the metric of the best-performing model in bold. See tables \ref{tab:odds_f1-1} and \ref{tab:odds_f1-2} in appendix \ref{appendix:add_rez} for the details of the obtained metrics.}
              \label{fig:bench_results}
        \end{figure*}
        
        Notice that for DROCC \cite{goyalDROCCDeepRobust2020}, GOAD \cite{goad}, and NeuTraL-AD \cite{neutralad}, we report in tables \ref{tab:odds_f1-1} and \ref{tab:odds_f1-2} the architecture that obtained the highest mean F1-score. The metrics obtained for the other architectures are detailed in tables \ref{tab:drocc}, \ref{tab:goad}, and \ref{tab:neutralad} in appendix \ref{appendix:add_rez}.
        We included each architecture of each approach to compute the mean rank as shown in \ref{fig:bench_results}. Following the literature, we report the average metrics over 20 runs. Our model was trained for each dataset on 4 or 8 Nvidia GPUs V100 16Go/32Go, depending on the dataset dimension. Note that the model can also be trained on a single GPU for small and medium datasets.
        
        For each dataset, we considered the same NPT architecture composed of $4$ layers alternating between Attention Between Datapoints and Attention Between Attributes and $4$ attention heads. Per \cite{kossen2021self}, we consider a Row-wise feed-forward (rFF) network with one hidden layer, 4x expansion factor, GeLU activation, and also include dropout with $p=0.1$ for both attention weights and hidden layers. We used LAMB \cite{lamb} with  $\beta = (0.9, 0.999)$ as the optimizer and also included a Lookahead \cite{lookahead} wrapper with slow update rate $\alpha = 0.5$ and $k = 6$ steps between updates. Similarly, following \cite{kossen2021self}, we consider a flat-then-anneal learning rate schedule: flat at the base learning rate for 70\% of steps and then anneals following a cosine schedule to 0 by the end of the training phase, and set gradient clipping at 1. 
        We chose $r$ in accordance with the masking probability $p_{mask}$ used during training and the total number of features $d$. We hypothesized that a too-high value of $r$ for a low $p_{mask}$ would pollute the anomaly score with reconstructions too challenging for the model, leading to high reconstruction error for both \textit{normal} samples and anomalies. Moreover, the hardest reconstructions, \textit{i.e.} those with a high number of masked features, would constitute a too high share of the total masks. Indeed, for a fixed $d$, $\binom{d}{k}$ as a function of $k$ is non-decreasing for $k\leq d/2$ and has an exponential growth rate. Furthermore, raising the value of the parameter $r$ can lead to a substantial augmentation in the number of masks $m$, consequently inducing a significant upsurge in the inference runtime.
        
        We detail in appendix \ref{appendix:exp_settings} the varying hyperparameters used for each dataset in our experiments. Notice that for most datasets, the hyperparameters remain unchanged. Variations of the hyperparameters are motivated by a swifter convergence of the training loss or computational costs for larger datasets. Each experiment can be replicated using the code made available on GitHub\footnote{\href{https://github.com/hugothimonier/NPT-AD/}{{https://github.com/hugothimonier/NPT-AD/}}}. 
        
        \paragraph{Results}
        As seen in figure \ref{fig:bench_results} and tables \ref{tab:odds_f1-1} and \ref{tab:odds_f1-2}, our model surpasses existing methods on most datasets by a significant margin regarding the F1-score. Moreover, our approach displays the highest mean F1-score and lowest mean rank over all datasets out of the $18$ tested approaches. The method of \citet{shenkar2022anomaly} ranks as the second highest in terms of average F1-score and the second highest mean rank over all datasets, while the simple KNN-based method \cite{ramaswamy2000efficient} ranks as the third best approach in terms of average F1-score and rank.  
        Also, our approach displays a smaller variance than competing methods except for COPOD \cite{Li2020}, which performs significantly worse than our approach regarding the F1-score and AUROC. The smaller variance could originate from the fact that our model uses, in a non-parametric fashion, the training set in the inference phase. This contributes to flattening the variations in the anomaly score attributed to discrepancies in the model's weights between runs.
        We also display in tables \ref{tab:odds_auc} and \ref{tab:odds_auc_2} in appendix \ref{appendix:add_rez} the AUROC for the same experiments and observe that we obtain the highest mean AUROC and the lowest mean rank while also displaying a smaller variance than other tested approaches.

    \section{Discussion}
    \label{sec:discussion}

        \begin{figure*}[ht!]
            \centering
            
            \subfigure[F1-score ($\uparrow$)]{
            \includegraphics[width=0.46\linewidth]{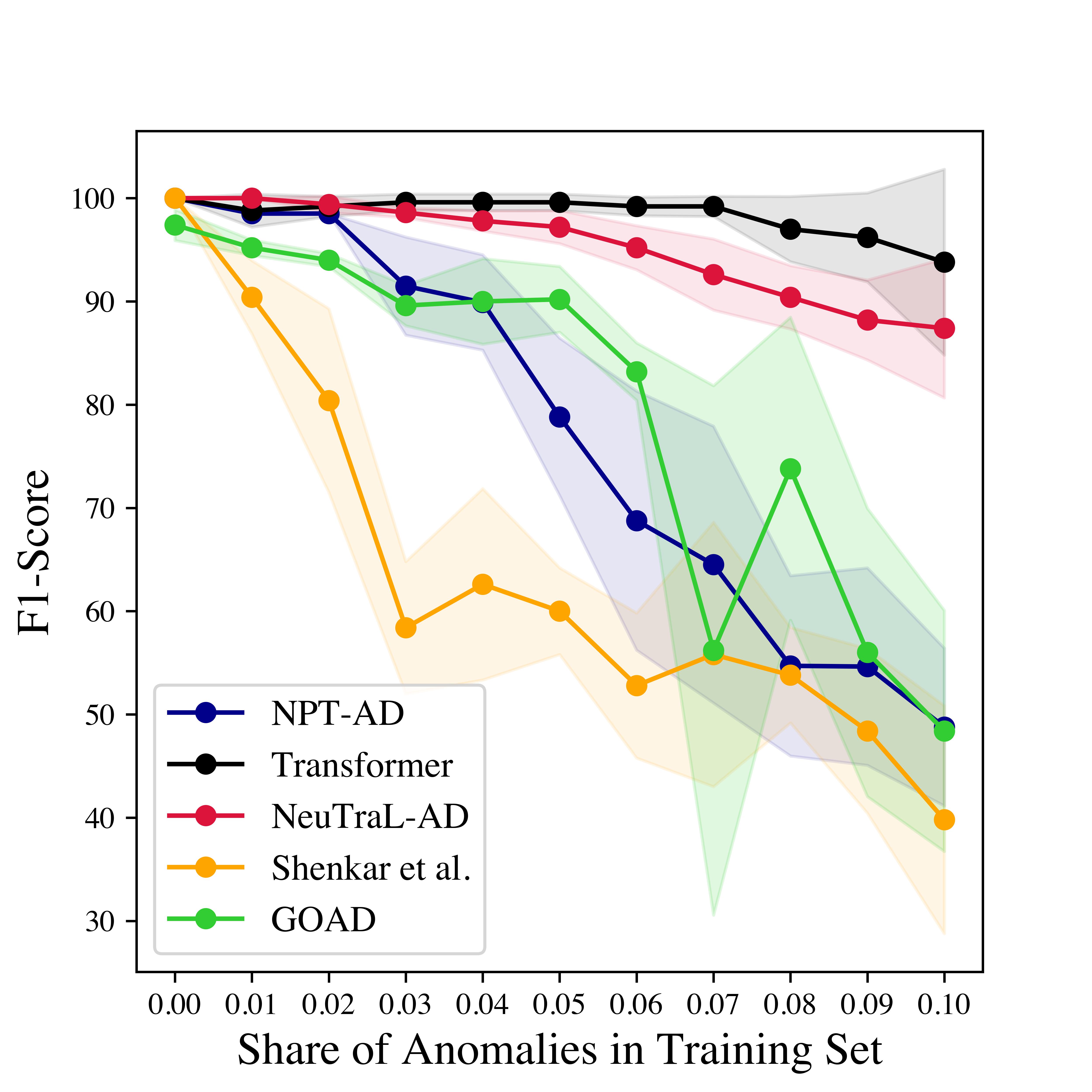}
            \label{fig:sub_F1}
            }
            \hfill
            \subfigure[AUROC ($\uparrow$)]{
            \includegraphics[width=0.46\linewidth]{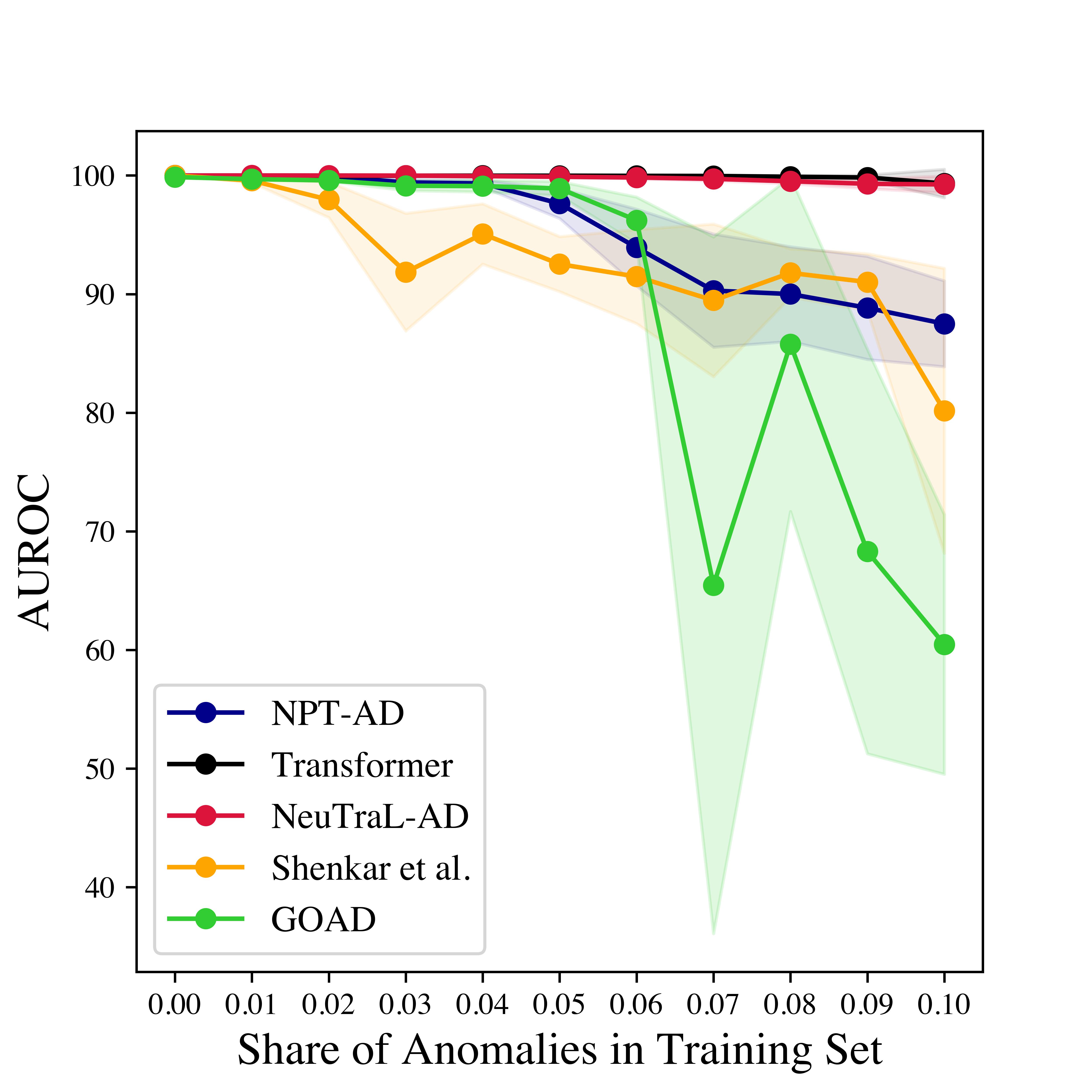}
            \label{fig:sub_AUROC}
            }
            
            \caption{Training set contamination impact on the F1-score and AUROC. Each model was trained $5$ times for each contamination share. The architecture used for NPT-AD is the same as for all experiments (see section \ref{sec:experiments}). The NPT and Transformer were trained for $100$ epochs with batch size equal to the dataset size, with learning rate $0.01$, optimizer LAMB \cite{lamb} with  $\beta = (0.9, 0.999)$, per-feature embedding dimension $16$, $r$ set to 1, and masking probability $p_{mask}=0.15$. NeuTraL-AD and GOAD were trained with hyperparameters as for the thyroid dataset in the original papers and \cite{shenkar2022anomaly} with its default parameters in their implementation.}
            \label{fig:contamination}
        \end{figure*}

        \subsection{Training set contamination} 
            Real-life anomaly detection applications often involve contaminated training sets; anomaly detection models must, therefore, be robust to small levels of dataset contamination. We experimented using a synthetic dataset to evaluate how much NPT-AD suffers from dataset contamination compared to recent deep AD methods. We constructed a synthetic dataset using two perfectly separable distributions for \textit{normal} and anomaly samples. Our training set contained $900$ \textit{normal} samples, and we kept aside $100$ anomaly samples that we could add to the training set. We considered $11$ different training sets with contamination shares ranging from $0$\% to $10$\% with a $1$\% step while keeping the validation set constant with a fixed composition of $10$\% anomalies and $90$\% \textit{normal} samples. We display the results of this experiment in Figure \ref{fig:contamination} in which we show how the performance of NPT-AD varies when the contamination share increases in comparison with a vanilla transformer \cite{attentionisallyouneed} for mask reconstruction, NeuTraL-AD \cite{neutralad}, GOAD \cite{goad} and the internal contrastive approach of \cite{shenkar2022anomaly}. 
            
            Our experimental results show that, as expected, the performance of NPT-AD deteriorates as the proportion of anomalies in the training set increases. For contamination shares lower than $2$\% (resp. $5$\%), the F1-score (resp. AUROC) remains close to its maximum value of $100$\%. However, the F1-score and AUROC deteriorate significantly for higher contamination levels while displaying a higher standard deviation. When anomalies constitute $10$\% of the training set, our approach achieves an average F1-score slightly lower than $50$\% and an average AUROC of $87$\%. 
            We observe that NPT-AD suffers less from dataset contamination than \cite{shenkar2022anomaly} for both F1-score and AUROC. We also notice that the method of \citet{shenkar2022anomaly} is particularly sensible to dataset contamination regarding the F1-score compared with NeuTraL-AD, GOAD, the transformer approach, and NPT-AD even for low contamination shares.
        
            Finally, we observe that the transformer approach suffers significantly less from training set contamination than NPT-AD, especially for high contamination shares. What could account for such a difference is the fact that the contamination effect is double for NPT-AD. First, during training, the model acquires the ability to reconstruct not only normal samples but also anomalies, thereby compromising the reconstruction error as a good proxy for anomalousness. Second, in inference, the model relies on the unmasked training set to reconstruct samples: normal samples may attend to anomaly samples, which may hamper the ability of the model to reconstruct them properly. Conversely, anomalies can attend to other anomalies, which helps reduce the reconstruction error for this class.
        
        \subsection{Sample-sample dependencies ablation study}
        \label{subsec:sample-sample}

            \begin{table}[ht!]
                \begin{center}
                    \caption{Ablation study. Comparison in AD performance between the vanilla transformer, Mask-KNN, and NPT-AD.}
                    \label{tab:compar_vanilla_trans}
                    \begin{tabular}{cccc}
                    \toprule
                          & Transformer & Mask-KNN & NPT-AD \\
                        \midrule
                         F1 & $57.4$ & $57.5$ & $\mathbf{68.8}$ \\
                         AUROC & $83.0$ & $84.5$ & $\mathbf{89.8}$ \\
                         \bottomrule
                    \end{tabular}
                \end{center}
            \end{table}
            To further explore the combined impact of sample-sample and feature-feature dependencies, we introduce a reconstruction-based technique similar to NPT-AD. We put forward Mask-KNN that relies on KNN imputation to reconstruct masked features (see alg. \ref{alg:mask-nn} in appendix \ref{app:mask_knn}). We also compare it to the vanilla transformer model trained in a framework similar to NPT-AD. Mask-KNN (resp. the transformer) can be seen as approximately equivalent to NPT-AD without considering the feature-feature dependencies (resp. the sample-sample dependencies) as illustrated in figure \ref{fig:dependencies} in appendix \ref{app:mask_knn}. Our experiment is detailed in appendix \ref{app:mask_knn} and summarized in tables \ref{tab:compar_vanilla_trans}, \ref{tab:odds1+mask-knn} and \ref{tab:mean_rk_mask_knn}. We observe that, indeed, NPT-AD outperforms the vanilla transformer and Mask-KNN based on both AUROC and F1-score as shown in table \ref{tab:compar_vanilla_trans}, emphasizing that combining both types of dependencies boosts anomaly detection performance.
            
            Furthermore, we anticipate observing distinct performance dynamics in NPT-AD compared to both the vanilla transformer and Mask-KNN. In scenarios where the identification of anomalies relies heavily on feature-feature dependencies, such as datasets where the vanilla transformer significantly outperforms Mask-KNN, we anticipate NPT-AD to surpass Mask-KNN. This superiority is attributed to NPT-AD's capacity to effectively harness feature-feature dependencies. Conversely, in datasets where sample-sample dependencies play a pivotal in detecting anomalies, \textit{i.e.} datasets on which Mask-KNN outperforms the vanilla transformer, we expect NPT-AD to surpass the vanilla transformer. This relationship between the performance metrics across the three approaches would validate that NPT-AD effectively integrates both feature-feature and sample-sample dependencies.
        
            As displayed in table \ref{tab:odds1+mask-knn}, our experiments show that we do observe this performance behavior. For instance, on datasets where Mask-KNN significantly outperforms the transformer by a sizable gap, \textit{e.g.} glass, vertebral, or vowels, NPT-AD also outperforms the transformer significantly. Similarly, on datasets where the transformer outperforms Mask-KNN, \textit{e.g.} lymphography, WBC or Forest Cover, NPT-AD also performs significantly better than Mask-KNN. This performance relation can help identify which datasets require which type of dependency to flag anomalies using reconstruction-based methods correctly. We propose such classification in table \ref{tab:feature_or_sample} in appendix \ref{app:mask_knn}.

    \section{Limitations and Conclusion}
    \label{sec:conclusion}

        \paragraph{Limitations} As with most non-parametric models, 
        NPT-AD displays a higher complexity than parametric approaches. NPT-AD can scale well for datasets with a reasonable number of features $d$; however, for large values of $d$, our approach involves a high computational cost in terms of memory and time. This cost originates from the complexity of NPT itself and how the anomaly score is derived. In table \ref{tab:runtime} in appendix \ref{appendix:comput_time}, we observe that NPT-AD displays longer runtime for datasets with large values of $d$ when $n$ is also high, \textit{e.g.} Mnist or backdoor. Two factors can account for this. First, the number of reconstructions highly depends on $d$, which increases the inference runtime; secondly, due to the feature embeddings, the dimension of the model also increases rapidly with $d$.
    
        \paragraph{Conclusion}
        In this work, we have proposed a novel deep anomaly detection method designed explicitly for tabular datasets. To the best of our knowledge,  our approach is the first to successfully utilize both feature-feature and sample-sample dependencies to identify anomalies. 
        Our experiments on an extensive benchmark demonstrate the effectiveness of our approach, outperforming existing state-of-the-art methods in terms of F1-score and AUROC. Our experiments further demonstrate the robustness of our method to a small training set contamination. Finally, this work emphasizes the importance of leveraging both dependencies to effectively detect anomalies on tabular datasets. 
        
        \paragraph{Future Work} Our work invites further research on the optimal way to combine sample-sample and feature-feature dependencies for anomaly detection on tabular data. In the current work, we relied on NPTs, which combine both dependencies \textit{internally}. However, this restrains pretext tasks used for self-supervised AD to tasks that NPTs can perform. Other forms of \textit{external} retrieval methods may be interesting to investigate to be able to \textit{augment} existing AD methods.

    \paragraph{Acknowledgment}
    This work was granted access to the HPC resources of IDRIS under the allocation 2023-101424 made by GENCI.
    This research publication is supported by the Chair "Artificial intelligence applied to credit card fraud detection and automated trading" led by CentraleSupelec and sponsored by the LUSIS company.
    \\ The authors would also like to thank Gabriel Kasmi for his helpful advice and feedback.
    
\bibliography{bibliography}
\bibliographystyle{icml2024}
\clearpage
\begin{appendices}

    \onecolumn

    \section{Multi-Head Self-Attention}
        \label{appendix:mhsa}
            Scaled dot-product attention  as first proposed in \cite{attentionisallyouneed} describes a mapping between queries $Q_i \in \mathbb{R}^{1\times h_k}$, keys $K_i \in \mathbb{R}^{1\times h_k}$ and values $V_i \in \mathbb{R}^{1\times h_v}$ to an output. The output is computed as a weighted sum of the values, where each weight is obtained by measuring the compatibility between queries and keys. 
            Take $\mathbf{Q} \in \mathbb{R}^{n\times h_k}$, $\mathbf{K} \in \mathbb{R}^{m\times h_k}$ and $\mathbf{V} \in \mathbb{R}^{m\times h_v}$ the corresponding matrices in which queries, keys, and values are stacked. Scaled dot-product attention is computed as
            \begin{equation}
                \mbox{Attention}(\mathbf{Q},\mathbf{K},\mathbf{V}) = \mbox{softmax}\left(\frac{\mathbf{Q}\mathbf{K}^\top}{\sqrt{h_k}}\right)\mathbf{V}
            \end{equation}
            where, for convenience, one often sets $h_k=h_v=h$.
            
            To foster the ability of a model to produce diverse and powerful representations of data samples, one often includes several dot-product attention mechanisms. Multi-head dot-product attention then describes the concatenation of $k$ independent attention heads:
            \begin{eqnarray}
                \mbox{MultiHead}(\mathbf{Q},\mathbf{K},\mathbf{V}) = \underset{\mbox{\tiny axis=k}}{\mbox{concat}}(O_1,\dots,O_k)W^O\\
                \mbox{where, }O_j = \mbox{Attention}(\mathbf{Q}W_j^Q,\mathbf{K}W_j^K,\mathbf{V}W_j^V)
            \end{eqnarray}
            where the embedding matrices $W_j^Q, W_j^K, W_j^V \in \mathbb{R}^{h\times h/k}$ are learned for each attention head $j \in \{1,\dots,k\}$ and $W^O \in \mathbb{R}^{h \times h}$ serves to mix the $h$ attention heads outputs.
            NPTs only include multi-head \textit{self}-attention mechanisms which consist in multi-head dot-product attention where queries, keys, and values are identical:
            \begin{equation}
                \mbox{MHSelfAtt}(\mathbf{H}) = \mbox{MultiHead}(\mathbf{Q}=\mathbf{H},\mathbf{K}=\mathbf{H},\mathbf{V}=\mathbf{H})
            \end{equation}
            As described in \cite{kossen2021self}, NPT follows transformer best practices to improve performances and involves a residual branch as well as layer normalization (LN) \cite{layernorm} before MHSelfAtt$(.)$. 
            \begin{equation}
                \mbox{Res}(\mathbf{H}) = \mathbf{H} W^{\mbox{\tiny res}} + \mbox{MHSelfAtt}\left(\mbox{LN}(\mathbf{H})\right)
            \end{equation}
            where $W^{\mbox{\tiny res}} \in \mathbb{R}^{h\times h}$ are learned weights. Layer normalization is also added after the residual branch as well as a row-wised feed-forward network (rFF):
            \begin{equation}
            \label{eq:mhsa}
                \mbox{MHSA}(\mathbf{H}) = \mbox{Res}(\mathbf{H}) + \mbox{rFF}\left(\mbox{LN}\left(\mbox{Res}\left(\mathbf{H}\right) \right) \right) \in \mathbb{R}^{n \times h}
            \end{equation}

    \section{Training Pipeline}
        \label{app:training_pipeline}

        Let $\mathbf{x} \in \mathcal{X} \subseteq \mathbb{R}^d$ be a sample with $d$ features, which can be either numerical or categorical. Let $e$ designate the hidden dimension of the transformer. The training pipeline consists of the following steps:
        \paragraph{Masking} We sample from a Bernouilli distribution with probability $p_{mask}$ whether each of the $d$ features is masked.
            $$
            \mbox{mask} = (m_1, \dots, m_{d}), 
            $$
        where $m_j \sim \mathcal{B}(1,p_{mask})$ $\forall j \in [1,...,d]$ and $m_j=1$ if feature $j$ is masked.

        \paragraph{Encoding} For numerical features, we normalize to obtain $0$ mean and unit variance, while we use one-hot encoding for categorical features. At this point, each feature $j$ for $j \in [1,2,...,d]$ has an $e_j$-dimensional representation, $encoded(\mathbf{x}_j) \in \mathbb{R}^{e_j}$ , where $e_j=1$ for numerical features and for categorical features $e_j$ corresponds to its cardinality. We then mask each feature according to the sampled mask vector and concatenate each feature representation with the corresponding mask indicator function. Hence, each feature $j$ has an $(e_j+1)$-dimensional representation 
        $$
        ((1 - m_j) \cdot encoded(\mathbf{x}_j), m_j) \in \mathbb{R}^{e_j+1},
        $$
        where $\mathbf{x}_j$ is the $j$-th features of sample $\mathbf{x}$.

        \paragraph{In-Embedding} We pass each of the features encoded representations of sample $\mathbf{x}$ through learned linear layers $\texttt{Linear}(e_j+1,e)$. We also learn $e$-dimensional index and feature-type embeddings as proposed in \cite{kossen2021self}. Both are added to the embedded representation of sample $\mathbf{x}$. The obtained embedded representation is thus of dimension $d \times e$
        $$
        e_{\mathbf{x}} = (e_{\mathbf{x}}^1,e_{\mathbf{x}}^2,\dots,e_{\mathbf{x}}^d) \in \mathbb{R}^{d \times e},
        $$
        and $e_{\mathbf{x}}^j \in \mathbb{R}^e$ is the embedded representation of feature $j$ for sample $\mathbf{x}$.

        \paragraph{NPT Encoder} We concatenate the embedded representations of $n$ samples and construct a matrix $\mathbf{X}$,
        $$
        \mathbf{X} = (e_{\mathbf{x}_1},\dots,e_{\mathbf{x}_n})^\top \in \mathbb{R}^{n\times d \times e} 
        $$
        that is fed to the NPT. Let $\mathbf{H} \in \mathbb{R}^{n\times d \times e}$ denote the output of the NPT,
        $$
        \mathbf{H} = \text{NPT}(\mathbf{X}) \in \mathbb{R}^{n\times d \times e}
        $$

        \paragraph{Out-Embedding} The output of the NPT, $\mathbf{H} \in \mathbb{R}^{n \times d\times e}$ is then used to compute an estimation of the original $d$-dimensional vector for each of the $n$ samples. To obtain the estimated feature $j$ for the $n$ samples, we take the $j$-th $n\times e$-dimensional representation which is output by the NPT encoder and pass it through a learned linear layer $\texttt{Linear}(e, e_j)$, where $e_j$ is $1$ for numerical features and the cardinality for categorical features. In other words, for each element $j$ in the second dimension of $\mathbf{H}$, $\mathbf{H}_{:,j:}$, we compute 
        $$
        \mathbf{Z} = \mathbf{H}_{:,j:} \mathbf{W}_{out}^{j} \in \mathbb{R}^{n\times e_j},
        $$
        where $\mathbf{W}_{out}^{j} \in \mathbb{R}^{e\times e_j}$, where $e_j=1$ for numerical features and for categorical features $e_j$ corresponds to its cardinality. For categorical feature, we take the $\arg\max$ along the second dimension of $\mathbf{Z}$.
                
    \clearpage
    \section{Datasets characteristics and experimental settings}
    \label{appendix:exp_settings}

        In table~\ref{tab:characteristics}, we display the main characteristics of the datasets involved in our experiments, while in \ref{tab:exp} we display the hyperparameters of our experiments.

        \begin{table}[h!]
        \caption{Datasets characteristics}
        \label{tab:characteristics}
        \begin{center}
            \begin{tabular}{cccc}
                \toprule
                Dataset &  $n$ & $d$ & Outliers \\
                \midrule
                
                Wine 
                &$129$ 
                &$13$
                & $10$ ($7.7$\%) \\
                
                Lympho
                & $148$
                & $18$
                & $6$ ($4.1$\%)\\
                
                Glass 
                &$214$ 
                &$9$
                & $9$ ($4.2$\%)\\
                
                Vertebral
                & $240$
                & $6$
                & $30$ ($12.5$\%)\\
                
                WBC 
                &$278$
                & $30$
                & $21$ ($5.6$\%)\\
                
                Ecoli
                & $336$ 
                &$7$
                & $9$ ($2.6$\%)\\
                
                Ionosphere
                & $351$
                & $33$ 
                &$126$ ($36$\%)\\
                
                Arrhythmia
                & $452$
                & $274$ 
                &$66$ ($15$\%)\\
                
                BreastW
                & $683$
                & $9$ 
                &$239$ ($35$\%)\\
                
                Pima
                & $768$
                & $8$
                & $268$ ($35$\%)\\
                
                Vowels
                & $1456$
                & $12$
                & $50$ ($3.4$\%)\\
                
                Letter Recognition
                & $1600$
                & $32$
                & $100$ ($6.25$\%)\\
                
                Cardio
                & $1831$
                & $21$ 
                &$176$ ($9.6$\%)\\
                
                Seismic
                & $2584$
                & $11$
                & $170$ ($6.5$\%)\\
                
                Musk
                & $3062$
                & $166$
                & $97$ ($3.2$\%)\\
                
                Speech
                & $3686$
                & $400$
                & $61$ ($1.65$\%)\\
                
                Thyroid
                & $3772$
                & $6$
                & $93$ ($2.5$\%)\\
                
                Abalone
                & $4177$
                & $9$
                & $29$ ($0.69$\%)\\
                
                Optdigits
                & $5216$
                & $64$
                & $150$ ($3$\%)\\
                
                Satimage-2
                & $5803$
                & $36$
                & $71$ ($1.2$\%)\\
                
                Satellite
                & $6435$
                & $36$
                & $2036$ ($32$\%)\\
                
                Pendigits 
                &$6870$ 
                &$16$
                & $156$ ($2.27$\%)\\
                
                Annthyroid
                & $7200$
                & $6$
                & $534$ ($7.42$\%)\\
                
                Mnist 
                &$7603$
                & $100$
                & $700$ ($9.2$\%)\\
                
                Mammography
                & $11183$
                & $6$
                & $260$ ($2.32$\%)\\
                
                Shuttle
                & $49097$
                & $9$ 
                &$3511$ ($7$\%)\\
                                        
                Mulcross
                & $262144$
                & $4$
                & $26214$ ($10$\%)\\
                
                ForestCover
                & $286048$
                & $10$
                & $2747$ ($0.9$\%)\\
                                        
                Campaign
                & $41188$
                & $62$ 
                & $4640$ ($11.3$\%) \\
                
                Fraud 
                & $284807$ 
                & $29$ 
                & $492$ ($0.17$\%) 
                \\

                Backdoor 
                & $95329$ 
                & $196$ 
                & $2329$ ($2.44$\%) 
                \\
                \bottomrule
            \end{tabular}
        \end{center}
    \end{table}
    
        \clearpage

        \begin{table}[h!]
        \caption{Datasets hyperparameters. When the batch size is $-1$ it refers to a full pass over the training set before an update of the weights.}
        \label{tab:exp}
        \begin{center}
            \begin{tabular}{ccccccccc}
                \toprule
                Dataset &  epoch & batch size & lr & $p^{train}_{mask}$ & $r$ & $m$ & $e$ \\
                \midrule
                Wine &$1000$ &$-1$& $0.001$ & $0.15$ & $1$ & $13$ & $8$ \\
                Lympho& $100$& $-1$& $0.01$ & $0.15$ & $4$ & $3078$ & $16$ \\
                Glass & $1000$& $-1$& $0.01$ & $0.15$ & $4$ & $255$ & $16$ \\
                Vertebral& $2000$& $-1$& $0.001$ & $0.15$ & $1$ & $6$ & $8$\\
                WBC &$100$& $-1$& $0.01$ & $0.15$ & $3$ & $4525$ & $16$\\
                Ecoli &$100$& $-1$& $0.01$ & $0.15$ & $3$ & $63$ & $16$\\
                Ionosphere &$100$& $-1$& $0.001$ & $0.15$ & $2$ & $561$ & $16$ \\
                Arrhythmia& $100$& $-1$& $0.01$ & $0.15$ & $1$ & $274$ & $16$ \\
                BreastW & $500$& $-1$& $0.01$ & $0.15$ & $3$ & $129$ & $16$\\
                Pima& $500$& $-1$& $0.01$ & $0.15$ & $4$ & $162$ & $16$ \\
                Vowels& $1000$& $-1$& $0.01$ & $0.15$ & $2$ & $78$ & $16$\\
                Letter Recognition & $1000$& $-1$& $0.01$ & $0.15$ & $1$ & $32$ & $16$\\
                Cardio& $100$ & $-1$& $0.01$ & $0.15$ & $2$ & $231$ & $16$ \\
                Seismic& $100$ & $-1$& $0.01$ & $0.15$ & $2$ & $276$ & $16$\\
                Musk& $100$ & $-1$& $0.01$ & $0.15$ & $2$ & $166$ & $16$\\
                Speech& $1000$ & $512$& $0.001$ & $0.15$ & $1$ & $400$ & $8$\\
                Thyroid& $5000$ & $-1$& $0.01$ & $0.1$ & $2$ & $21$ & $16$\\
                Abalone& $1000$ & $-1$& $0.0001$ & $0.15$ & $4$ & $162$ & $16$\\
                Optdigits& $500$ & $-1$& $0.01$ & $0.2$ & $1$ & $64$ & $16$\\
                Satimage-2& $100$ & $-1$& $0.01$ & $0.2$ & $1$ & $36$ & $16$\\
                Satellite& $100$ & $-1$& $0.01$ & $0.2$ & $1$ & $36$ & $16$ \\
                Pendigits & $1000$ & $-1$& $0.01$ & $0.25$ & $2$ & $136$ & $16$ \\
                Annthyroid& $400$ & $-1$& $0.01$ & $0.15$ & $1$ & $6$ & $16$\\
                Mnist & $1000$ & $-1$& $0.001$ & $0.15$ & $1$ & $100$ & $32$ \\
                Mammography& $200$& $-1$& $0.01$ & $0.25$ & $4$ & $56$ & $16$\\
                Shuttle& $100$& $4096$& $0.01$ & $0.25$ & $3$ & $129$ & $64$\\
                Mulcross& $100$& $4096$& $0.001$ & $0.15$ & $2$ & $10$ & $16$\\
                ForestCover& $100$& $4096$& $0.01$ & $0.15$ & $2$ & $55$ & $16$\\
                Campaign & $100$ & $4096$ & $0.001$ & $0.15$ & $1$ & $62$ & $16$ \\
                Fraud & $100$ & $4096$ & $0.001$ & $0.2$ & $1$ & $29$ & $32$ \\
                Backdoor & $1000$ & $256$ & $0.001$ & $0.2$ & $1$ & $196$ & $32$ \\
                \bottomrule
            \end{tabular}
        \end{center}
    \end{table}
    Hyperparameter selection was done so as to obtain the fastest training loss convergence. The pipeline consisted of the following:
    \begin{itemize}
        \item \textbf{Hidden dimension} $e$: We started from smaller to larger models to achieve convergence.
        \item \textbf{Batch size}: We maximized batch size to fit into memory. Larger batch sizes are beneficial to our approach since a larger number of samples in the same batch increases the samples to which each sample can attend and fosters better learning of sample-sample dependencies.
        \item \textbf{Learning rate} (lr) was selected to achieve the fastest loss convergence for each architecture.
        \item \textbf{Masking probability} $p_{mask}$: we started from $0.25$ and reduced with a step $0.05$ until loss convergence.
        \item  \textbf{Maximum number of features masked simultaneously } $r$: it was chosen in accordance with the chosen $p_{mask}$ and the overall number of feature $d$. We also considered inference time when setting $r$ as setting $r$ higher than $1$ for a dataset with many features would make inference time explode. Take the Speech dataset with $d=400$; going from $r=1$ to $r=2$ would make the number of masks go from $400$ to $82200$. 
    \end{itemize}
        \clearpage
        
    \section{Additional experiments}
    \label{appendix:results}

        \subsection{Additional results}
            \label{appendix:add_rez}
            In this section, we display the metrics for each of the experiments we performed. This includes the F1-score for all tested approaches in tables \ref{tab:odds_f1-1} and \ref{tab:odds_f1-2}, the AUROC for the approaches for which it is relevant to compute it; displayed in tables \ref{tab:odds_auc} and \ref{tab:odds_auc_2}, the F1-score for each architecture discussed in the original papers of NeuTraL-AD \cite{neutralad} in table \ref{tab:neutralad}, GOAD \cite{goad} in table \ref{tab:goad}, and DROCC \cite{goyalDROCCDeepRobust2020} in table \ref{tab:drocc}. For each of these tables, we highlight in bold the highest metric in the table.

            \begin{table*}[h!]
                \begin{center}
                    \caption{Anomaly detection F1-score ($\uparrow$) for deep models. We perform $5$\% T-test to test whether the difference between the highest metrics for each dataset is statistically significant.}
                    \label{tab:odds_f1-1}
\begin{tabular}{lccccccc}
                            \toprule
                            Method 
                            & DROCC 
                            & GOAD 
                            & NeuTraL.
                            & Inter.Cont. 
                            & Transformer
                            & \mbox{NPT-AD} \\
                            & (abalone) & (thyroid) & (arrhy.) & & \\
                            \midrule
                            
                            Wine 
                            &$63.0\pm$\small{$20.0$}
                            &$67.0\pm$\small{$9.4$} 
                            &$78.2\pm$\small{$4.5$} 
                            &$90.0\pm$\small{$6.3$} 
                            &$23.5\pm$\small{$7.9$}
                            &$72.5\pm$\small{$7.7$}\\
                            
                            Lympho 
                            & $65.0\pm$\small{$5.0$}
                            & $68.3\pm$\small{$13.0$}  
                            & $20.0\pm$\small{$18.7$}  
                            & $86.7\pm$\small{$6.0$} 
                            & $88.3\pm$\small{$7.6$}
                            & $\mathbf{94.2}\pm$\small{$7.9$}\\
                            
                            Glass 
                            & $14.5$\small{$\pm11.1$}
                            & $12.7\pm$\small{$3.9$}
                            & $9.0\pm$\small{$4.4$}
                            & $\mathbf{27.2}$\small{$\pm10.6$}
                            & $14.4\pm$\small{$6.1$}
                            & $\mathbf{26.2}$\small{$\pm10.9$} \\
\bottomrule
                    \end{tabular}
                \end{center}
            \end{table*}

\clearpage\vspace*{0pt}
            \begin{table*}[t!]
                \begin{center}
                    \caption{Anomaly detection F1-score ($\uparrow$) for non-deep models. We perform $5$\% T-test to test whether the difference between the highest metrics for each dataset is statistically significant.}
                    \label{tab:odds_f1-2}
                    \begin{tabular}{lcccccc}
                        \toprule
                            Method  
                            & COPOD 
                            & IForest 
                            & KNN 
                            & PIDForest 
                            & RRCF
                            & \mbox{NPT-AD} \\
                        \midrule 
                            Wine
                            &$60.0\pm$\small{$4.5$}
                            &$64.0\pm$\small{$12.8$}
                            &$\mathbf{94.0}\pm$\small{$4.9$}
                            &$50.0\pm$\small{$6.4$}
                            &$69.0\pm$\small{$11.4$}
                            &$\mathbf{72.5}\pm$\small{$7.7$}\\
            
                            Lympho 
                            & $85.0\pm$\small{$5.0$}
                            & $71.7\pm$\small{$7.6$}
                            & $80.0\pm$\small{$11.7$}
                            & $70.0\pm$\small{$0.0$}
                            & $36.7\pm$\small{$18.0$}
                            & $\mathbf{94.2}\pm$\small{$7.9$}\\
            
                            Glass
                            & $11.1\pm$\small{$0.0$}
                            & $11.1\pm$\small{$0.0$}
                            & $11.1\pm$\small{$9.7$}
                            & $8.9\pm$\small{$6.0$}
                            & $15.6\pm$\small{$13.3$}
                            & $\mathbf{26.2}$\small{$\pm10.9$} \\
            
                            Vertebral
                            & $1.7\pm$\small{$1.7$}
                            & $13.0\pm$\small{$3.8$}
                            & $10.0\pm$\small{$4.5$}
                            & $12.0\pm$\small{$5.2$}
                            & $8.0\pm$\small{$4.8$}
                            & $\mathbf{20.3}\pm$\small{$4.8$}\\
\bottomrule

                    \end{tabular}
                \end{center}
            \end{table*}
\clearpage\vspace*{0pt}
            \begin{table*}[t!]
                \begin{center}
                    \caption{Anomaly detection AUROC($\uparrow$). We perform $5$\% T-test to test whether the differences between the highest metrics for each dataset are statistically significant.}
                    \label{tab:odds_auc}
                    \begin{tabular}{lcccccccc}
                       \toprule
                       Method & DROCC & DROCC & DROCC & GOAD & GOAD & GOAD  & Internal Cont. & \mbox{NPT-AD} \\
                       & \footnotesize{(Thyroid)} & \footnotesize{(Arrhyth.)} & \footnotesize{(Abalone)} &\footnotesize{(Thyroid)} &\footnotesize{(kddrev)} & \footnotesize{(kdd)} & & (Ours) \\
                       \midrule
                       
                        Wine 
                        & $53.5\pm$\small{$22$}
                        & $60.1\pm$\small{$32$}
                        & $90.9\pm$\small{$8.2$}
                        & $95.2\pm$\small{$1.9$}
                        & $97.3\pm$\small{$1.7$}
                        & $86.3\pm$\small{$9.5$}
                        & $\mathbf{99.5}\pm$\small{$0.6$}
                        & $96.6\pm $\small{$0.5$}\\
                        
                        Lympho 
                        & $6.4\pm$\small{$5.2$}
                        & $58.6\pm$\small{$30$}
                        & $83.7\pm$\small{$12$}
                        & $94.8\pm$\small{$5.6$}
                        & $79.7\pm$\small{$11$}
                        & $59.9\pm$\small{$15$}
                        & $99.5\pm$\small{$0.3$}
                        & $\mathbf{99.9}\pm$\small{$0.1$} \\
                        
                        Glass 
                        & $63.5\pm$\small{$9.1$}
                        & $55.5\pm$\small{$22$}
                        & $75.4\pm$\small{$8.9$}
                        & $62.2\pm$\small{$14$}
                        & $85.5\pm$\small{$7$}
                        & $82.1\pm$\small{$6.3$}
                        & $\mathbf{88.1}\pm$\small{$5.0$}
                        & $82.8\pm$\small{$2.4$}\\
\bottomrule
                    \end{tabular}
                \end{center}
            \end{table*}
\clearpage\vspace*{0pt}
            \begin{table*}[t!]
                \begin{center}
                    \caption{Anomaly detection AUROC($\uparrow$). We perform $5$\% T-test to test whether the differences between the highest metrics for each dataset are statistically significant.}
                    \label{tab:odds_auc_2}
                    \begin{tabular}{lcccccc}
                       \toprule
                        Method & NeuTraL-AD & NeuTraL-AD & NeuTraL-AD & NeuTraL-AD & COPOD & Transformer \\
                        &\footnotesize{(thyroid)} &\footnotesize{(arrhythmia)}  &\footnotesize{(kddrev)} & \footnotesize{(kdd)} \\
                       \midrule
                        Wine 
                        & $82.9\pm$\small{$13.1$} 
                        & $\mathbf{95.4}\pm$\small{$1.9$} 
                        & $86.0\pm$\small{$10.3$} 
                        & $85.2\pm$\small{$12.9$} 
                        & $87.5\pm$\small{$1.7$}
                        & $61.4\pm $\small{$6.7$}
                                    \\ 
                        Lympho 
                        & $90.7\pm$\small{$5.1$} 
                        & $80.9\pm$\small{$7.5$} 
                        & $93.1\pm$\small{$4.1$} 
                        & $83.1\pm$\small{$9.9$} 
                        & $\mathbf{99.4}\pm$\small{$0.4$}
                        & $\mathbf{99.5}\pm $\small{$0.4$}
                                    \\ 
                        Glass 
                        & $62.2\pm$\small{$3.0$} 
                        & $62.9\pm$\small{$1.2$} 
                        & $62.5\pm$\small{$4.6$} 
                        & $\mathbf{65.1}\pm$\small{$2.9$} 
                        & $63.7\pm$\small{$3.3$}
                        & $61.2\pm $\small{$7.0$}
                                    \\ 
                        Vertebral 
                        & $32.8\pm$\small{$3.5$} 
                        & $25.8\pm$\small{$4.5$} 
                        & $51.9\pm$\small{$14.9$} 
                        & $\mathbf{54.4}\pm$\small{$16.3$} 
                        & $32.6\pm$\small{$1.2$}
                        & $44.8\pm $\small{$5.2$}
                                    \\
\bottomrule
            \end{tabular}
           
                \end{center}
            \end{table*} 
            \clearpage\vspace*{0pt}
            \begin{table}[t!]
            \begin{center}
                \caption{DROCC \cite{goyalDROCCDeepRobust2020}: F1-score ($\uparrow$) between architecture. The mean rank was computed including all architectures of all models.}
                \label{tab:drocc}
                \begin{tabular}{lccc}
               \toprule
               Method & DROCC & DROCC & DROCC \\
               & \footnotesize{(thyroid)} & \footnotesize{(arrhyth.)} & \footnotesize{(abalone)}  \\
               \midrule
                Wine & $20.0\pm$\small{$19.0$} &$32.0\pm$\small{$35.4$} &$\mathbf{63.0}\pm$\small{$20.0$} \\
                
                Lympho & $0.0\pm$\small{$0.0$} & $38.3\pm$\small{$23.6$}& $\mathbf{65.0}\pm$\small{$5.0$}\\
                
                Glass & $\mathbf{22.2}$\small{$\pm17.2$}& $13.3$\small{$\pm12.0$}& $14.5$\small{$\pm11.1$} \\
                
                Vertebral &$25.7\pm$\small{$5.4$}& $\mathbf{27.0}\pm$\small{$15.9$}& $9.3$$\pm$\small{$6.1$} \\
                
                Wbc & $0.0\pm$\small{$0.0$}& $\mathbf{18.6}\pm$\small{$16.0$}& $9.0\pm$\small{$6.2$}\\ 
                
                Ecoli & N/A & N/A & N/A \\
\bottomrule
            \end{tabular}
            \end{center}
        \end{table} 
        
            \clearpage
 
            \begin{table}[h!]
            \begin{center}
                \caption{GOAD \cite{goad}: F1-score ($\uparrow$) between architecture. The mean rank was computed including all architectures of all models.}
                \label{tab:goad}
                \begin{tabular}{lccc}
               \toprule
               Method & GOAD & GOAD & GOAD \\
               &\footnotesize{(thyroid)} &\footnotesize{(kddrev)} & \footnotesize{(kdd)} \\
               \midrule
                Wine 
                &$67.0\pm$\small{$9.4$} 
                &$\mathbf{76.0}\pm$\small{$10.8$} 
                &$42.2\pm$\small{$26.9$} \\
                
                Lympho 
                & $\mathbf{68.3}\pm$\small{$13.0$} 
                & $67.7\pm$\small{$7.8$}
                & $46.0\pm$\small{$21.5$} \\
                
                Glass 
                & $12.7$\small{$\pm3.9$}
                & $\mathbf{25.7}$\small{$\pm12$}
                & $24.0$\small{$\pm15.1$} \\
                
                Vertebral 
                & $16.3\pm$\small{$9.6$}
                & $\mathbf{26.9}\pm$\small{$5.2$}
                & $25.5\pm$\small{$4.7$}\\
                
                Wbc 
                & $\mathbf{66.2}\pm$\small{$2.9$}
                & $16.8\pm$\small{$16.1$}
                & $57.2\pm$\small{$6.9$}\\ 
                
                Ecoli 
                & $61.4\pm$\small{$31.7$}
                & $\mathbf{69.3}\pm$\small{$23.7$}
                & $66.1\pm$\small{$27.8$} \\
\bottomrule
            \end{tabular}
            \end{center}
        \end{table}
        
            \clearpage\vspace*{0pt}
            \begin{table*}[t!]
            \begin{center}
                \caption{NeuTraL-AD \cite{neutralad}: F1-score ($\uparrow$) between architecture. The mean rank was computed including all architectures of all models.}
                 \label{tab:neutralad}
                \begin{tabular}{lcccc}
               \toprule
               Method & NeuTraL-AD & NeuTraL-AD & NeuTraL-AD & NeuTraL-AD\\
               &\footnotesize{(thyroid)} &\footnotesize{(arrhythmia)}  &\footnotesize{(kddrev)} & \footnotesize{(kdd)} \\
               \midrule
                Wine &$51.4\pm$\small{$26.2$} 
                & $\mathbf{78.2}\pm$\small{$4.5$}  
                & $62.3\pm$\small{$26.9$}  
                & $62.3\pm$\small{$28.9$} 
                \\
                
                Lympho & $46.7\pm$\small{$17.9$} 
                & $20\pm$\small{$18.7$}
                & $\mathbf{54.2}\pm$\small{$15.7$}  
                & $34.2\pm$\small{$15.3$} 
                \\
                
                Glass & $7.5$\small{$\pm6.2$}
                & $9.0\pm$\small{$4.4$}
                & $9.5\pm$\small{$5.9$} 
                &$\mathbf{13.0}\pm$\small{$7.8$} 
                \\
                
                Vertebral & $9.2\pm$\small{$3.4$}
                & $3.8\pm$\small{$1.2$} 
                & $\mathbf{23.3}\pm$\small{$9.8$} 
                &$13.0\pm$\small{$7.8$} 
                \\
                
                Wbc & $40.7\pm$\small{$10$}
                & $\mathbf{60.9}\pm$\small{$5.6$} 
                & $21.1\pm$\small{$10$} 
                &$13.0\pm$\small{$7.8$} 
                \\ 
                
                Ecoli & $4.0\pm$\small{$5.8$}
                & $7.0\pm$\small{$7.1$}
                & $6.5\pm$\small{$11.1$} 
                &$\mathbf{8.0}\pm$\small{$12.5$} 
                \\
\bottomrule
            \end{tabular}
            \end{center}
        \end{table*}
            \clearpage
    
        \subsection{Mask-KNN}
            \label{app:mask_knn}

    \begin{figure}[h!]
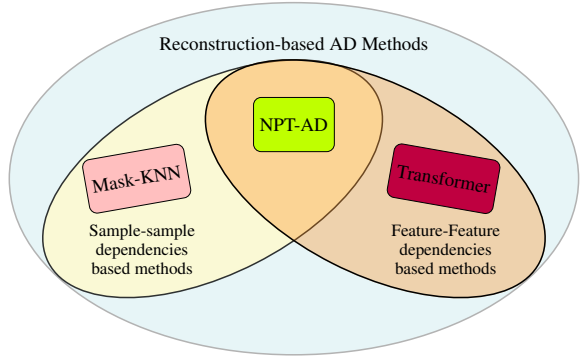

        \begin{center}
        \resizebox{0.45\textwidth}{!}{
            %
           }
           \captionof{figure}{While Mask-KNN only relies on sample-sample dependencies and the vanilla transformer only attends to feature-feature dependencies, NPT-AD combines both for anomaly detection.}
            \label{fig:dependencies}
        \end{center}
   \end{figure}

    To further investigate the impact of combining feature-feature and sample-sample dependencies, we rely on reconstruction-based strategy which makes use of the KNN-Imputer strategy and compare to the vanilla transformer and NPT-AD. 

    \paragraph{K-Nearest Neighbor Imputation}
    Take a dataset $\mathcal{D}=\{\mathbf{x}_i\}_{i=1}^n$ where $\mathbf{x}_i\in \mathbb{R}^d$ and for which some samples might display missing values in the feature vector. K-nearest neighbor imputation for a sample $\mathbf{z} \in \mathcal{D}$ consists in identifying the $k$ nearest neighbors of sample $\mathbf{z}$ given a distance measure $d:\mathbb{R}^d \times \mathbb{R}^d \to \mathbb{R}$, where $k$ is a hyperparameter that must be discretionary chosen. This distance measure only takes into account the non-missing features of sample $\mathbf{z}$. Let $\mathcal{I}$ designate the index of the non-missing features and $\mathbf{z}^{[\mathcal{I}]}$ the corresponding features of sample $\mathbf{z}$, then the $k$-nearest neighbors of sample $\mathbf{z}$ are identified through evaluating the distance $d(\mathbf{z}^{[\mathcal{I}]},\mathbf{x}_j^{[\mathcal{I}]})$ for each $\mathbf{x}_j \in \mathcal{D}$ and ordering them to find the $k$ smallest. Let $\mathcal{K}(z)$ designate the $k$ nearest neighbors of sample $\mathbf{z}$, $\bar{\mathcal{I}}$ the missing values of $\mathbf{z}$, then $\forall i \in \bar{\mathcal{I}}$
    \begin{equation}
        \hat{\mathbf{z}}^{i} = \frac{1}{k} \sum_{\mathbf{x} \in \mathcal{K}(\mathbf{z})} \mathbf{x}^{i}. 
    \end{equation}
    Other imputation methods include weighting each sample in $\mathcal{K}(\mathbf{z})$ by its inverse distance to $\mathbf{z}$, denoted $\omega_{(\mathbf{z}, \mathbf{x})}^{[\mathcal{I}]} = 1/d(\mathbf{z}^{[I]}, \mathbf{x}_j^{[\mathcal{I}]})$. This gives 
    \begin{equation}
        \label{eq:mask_knn_weighted}
        \hat{\mathbf{z}}^{i} = \frac{1}{\sum_{\mathbf{x}} \omega_{(\mathbf{z}, \mathbf{x})}^{[\mathcal{I}]}} \sum_{\mathbf{x} \in \mathcal{K}(\mathbf{z})} \omega_{(\mathbf{z}, \mathbf{x})}^{[\mathcal{I}]} \mathbf{x}^{i}. 
    \end{equation}
    This approach leverage only sample-sample dependencies, while the vanilla transformer only leverage feature-feature dependencies and NPT-AD combines these two types as illustrated in figure \ref{fig:dependencies}.

    \paragraph{Mask-KNN Anomaly Score}
    Consider a training set $\mathcal{D}_{train} = \{\mathbf{x}_i\}_{i=1}^{n_{train}}$, $\mathbf{x}_i \in \mathbb{R}^d$ comprised of only \textit{normal} samples and a validation set $\mathcal{D}_{val} = \{\mathbf{x}_i\}_{i=1}^{n_{val}}$ for which we wish to predict the label. In a reconstruction-based approach we construct an anomaly score based on how masked samples are well-reconstructed using KNN imputation as described in the previous paragraph. First, we construct a mask bank comprised of $m$ masks, where $m = \sum_{j=1}^r \binom{d}{j}$ and $r$ designates the maximum number of features masked simultaneously. The mask bank is comprised of all possible combinations of $j$ masked features for $j\leq r$. Each mask corresponds to a $d$-dimensional vector composed of $0$ and $1$, where $1$'s indicate that the corresponding features will be masked.
    Let us denote as $\hat{\mathbf{z}}^{(\ell)}$ the reconstructed sample $\mathbf{z}$ for mask $\ell$, take $d:\mathbb{R}^d \times \mathbb{R}^d \to \mathbb{R}$ a distance measure, \textit{e.g.} the $\ell_2$-norm, then the anomaly score for sample $\mathbf{z}$ is given as
    \begin{equation}
        \mbox{Mask-KNN}(\mathbf{z}) = \sum_{\ell=1}^m d(\mathbf{z},\hat{\mathbf{z}}^{(\ell)})
    \end{equation}
    We give the pseudo-code of this method in alg. \ref{alg:mask-nn}.

   \begin{algorithm}[tb]
   \caption{Pseudo Python Code for Mask-KNN}
   \label{alg:mask-nn}
    \begin{algorithmic}
   \STATE {\bfseries Require:} $
        \mathcal{D}_{train} \in \mathbb{R}^{n_{train}\times d}, \mathcal{D}_{val}\in \mathbb{R}^{n_{val}\times d}$, $k$, $mask\_bank$, $d:\mathbb{R}^d \times \mathbb{R}^d \to \mathbb{R}$
        
        \STATE $\mbox{Mask-KNN} \gets dict()$ 
        \STATE $B \gets$ random sample of size $b$ from $\mathcal{D}_{train}$
        \FOR{$\texttt{mask} \in mask\_bank$}
            \FOR{$\texttt{idx} \in \texttt{range}(n_{val})$}
                \STATE $\mathbf{z} \gets \mathcal{D}_{val}[\texttt{idx},:]$
                \STATE $\tilde{\mathbf{z}} \gets apply\_mask(\mathbf{z}, \texttt{mask})$
                \STATE $\mathbf{X} \gets (\tilde{\mathbf{z}}, B)^\top$
                \STATE $\hat{\mathbf{X}} \gets \mbox{KNNImputer}(\mathbf{X}, k)$
                \STATE $\hat{\mathbf{z}} \gets \hat{\mathbf{X}}[0,:]$
                \STATE $\mbox{Mask-KNN}[\texttt{idx}] \mathrel{{+}{=}}  d(\mathbf{z},\hat{\mathbf{z}})$
            \ENDFOR
        \ENDFOR
\end{algorithmic}
\end{algorithm}

    \paragraph{Implementation}
        For simplicity we set $r$ to $2$ for all experiments, except for large dataset ($n>200,000$) for which $r$ was set to $1$ for computational reasons. We set $k$, the number of neighbors, to $5$ as for the vanilla KNN implementation. When present, categorical features were encoded using one-hot encoding. Except for large datasets ($n>200,000$) with many features, $d$, such as ForestCover, Fraud and Backdoor, we set $B$ as the entire training set. Otherwise, we take a random subsample of size $b=10,000$. We use the imputation strategy described in equation \ref{eq:mask_knn_weighted} to reconstruct the masked sampled. We report the results of this experiment in table \ref{tab:odds1+mask-knn} and compare the performance of Mask-KNN to KNN, the internal contrastive approach of \cite{shenkar2022anomaly} and NPT-AD. We run the algorithm $20$ times for each dataset, except for ForestCover, Fraud and Backdoor, for which report an average over $10$ runs for computational reasons. The mean rank, provided in table \ref{tab:odds1+mask-knn}, was computed, including each architecture of each approach. For completeness, we also include a table containing the mean rank of all approaches including MasK-KNN in table \ref{tab:mean_rk_mask_knn}.
    
    \paragraph{Results}
        We observe that Mask-KNN obtains satisfactory results on a significant share of the tested datasets, \textit{e.g.} pendigits, satellite; while also displaying poor performance on some datasets such as forest or backdoor in comparison with NPT-AD. Several factors can account for this. First, NPTs automatically select the number of relevant samples on which to rely to reconstruct the masked features, thus making this approach much more flexible than Mask-KNN, which has a fixed number of neighbors. Second, NPT-AD relies on attention mechanisms to learn the weights attributed to relevant samples while Mask-KNN relies on the $\ell_2$-distance. Although the $\ell_2$-distance offers a precise measure of similarity based on geometric distance, the attention mechanism can capture much more complex relations between samples. Finally, NPT-AD not only relies on sample-sample dependencies to reconstruct the mask features, but it also attends to feature-feature dependencies.
        
        The strong performance of NPT-AD on datasets where Mask-KNN also performs well serves as evidence supporting the fact that NPT-AD effectively captures sample-sample dependencies. Moreover, NPT-AD outperforms Mask-KNN on most datasets where the approach of \cite{shenkar2022anomaly} performs well, highlighting the crucial role of feature-feature dependencies on specific datasets. 
        The results displayed in table \ref{tab:odds1+mask-knn} show that NPT-AD manages to capture both feature-feature and sample-sample dependencies to reconstruct samples when sample-sample dependencies are not sufficient.
    \clearpage
    \vspace*{0pt}
    \begin{table*}[t!]
        \begin{center}
            \caption{Anomaly detection F1-score ($\uparrow$). We perform $5$\% T-test to test whether the difference between the highest metrics for each dataset is statistically significant. \textbf{Apart from this table, Mask-KNN was not included in the computation of the mean rank.} The mean rank for the F1-score of all approaches including Mask-KNN is displayed in table \ref{tab:mean_rk_mask_knn}.}
            \label{tab:odds1+mask-knn}
            \begin{tabular}{lccc}
                \toprule
                Method 
                & Transformer
                & \mbox{NPT-AD} 
                & Mask-KNN \\
                \midrule
                
                Wine 
                & $23.5\pm$\small{$7.9$}
                & $\mathbf{72.5}\pm$\small{$7.7$}
                & $28.0\pm$\small{$18.1$}\\
                
                Lympho 
                & $88.3\pm$\small{$7.6$}
                & $\mathbf{94.2}\pm$\small{$7.9$}
                & $60.0\pm$\small{$12.2$} \\
                
                Glass 
                & $14.4\pm$\small{$6.1$}
                & $\mathbf{26.2}$\small{$\pm10.9$} 
                & $\mathbf{26.7}\pm$\small{$5.4$}\\
                
                Vertebral 
                & $12.3\pm$\small{$5.2$}
                & $20.3\pm$\small{$4.8$}
                & $\mathbf{24.7}\pm$\small{$5.9$}\\
                
                Wbc 
                & $66.4\pm$\small{$3.2$}
                & $\mathbf{67.3}\pm$\small{$1.7$}
                & $\mathbf{68.1}\pm$\small{$3.0$}\\ 
                
                Ecoli 
                & $75.0\pm$\small{$9.9$}
                & $\mathbf{77.7}\pm$\small{$0.1$}
                & $63.9\pm$\small{$6.9$} \\
                
                Ionosph. 
                & $88.1\pm$\small{$2.8$}
                & $\mathbf{92.7}\pm$\small{$0.6$}
                & $89.7\pm$\small{$0.9$}\\
                
                Arrhyth. 
                & $59.8\pm$\small{$2.2$}
                & $60.4\pm$\small{$1.4$}
                & $\mathbf{62.9}\pm$\small{$2.4$}\\
                
                Breastw 
                & $\mathbf{96.7}\pm$\small{$0.3$}
                & $95.7\pm$\small{$0.3$}
                & $96.2\pm$\small{$0.7$}\\
                
                Pima 
                & $65.6\pm$\small{$2.0$}
                & $\mathbf{68.8}\pm$\small{$0.6$}
                & $63.5\pm$\small{$1.8$}\\
                
                Vowels 
                & $28.7\pm$\small{$8.0$}
                & $\mathbf{88.7}\pm$\small{$1.6$}
                & $84.3\pm$\small{$4.9$}\\
                
                Letter 
                & $41.5\pm$\small{$6.2$}
                & $\mathbf{71.4}\pm$\small{$1.9$}
                & $56.7\pm$\small{$3.2$} \\
                
                Cardio 
                & $68.8\pm$\small{$2.8$}
                & $\mathbf{78.1}\pm$\small{$0.1$}
                & $69.7\pm$\small{$2.0$}\\
                
                Seismic 
                & $19.1\pm$\small{$5.7$}
                & $\mathbf{26.2}\pm$\small{$0.7$} 
                & $\mathbf{26.2}\pm$\small{$1.7$} \\
                
                Musk 
                & $\mathbf{100}\pm$\small{$0.0$}
                & $\mathbf{100}\pm$\small{$0.0$}
                & $\mathbf{100}\pm$\small{$0.0$}\\
                
                Speech 
                & $6.8\pm$\small{$1.9$}
                & $\mathbf{9.3}\pm$\small{$0.8$}
                & $\mathbf{10.2}\pm$\small{$2.9$}\\
                
                Thyroid 
                & $55.5\pm$\small{$4.8$}
                & $\mathbf{77.0}\pm$\small{$0.6$}
                & $31.6\pm$\small{$5.4$}\\
                
                Abalone 
                & $42.5\pm$\small{$7.8$}
                & $\mathbf{59.7}\pm$\small{$0.1$}
                & $43.2\pm$\small{$7.0$}\\
                
                Optdigits 
                & $61.1\pm$\small{$4.7$}
                & $62.0\pm$\small{$2.7$}
                & $\mathbf{89.0}\pm$\small{$1.0$}\\
                
                Satimage
                & $89.0\pm$\small{$4.1$}
                & $\mathbf{94.8}\pm$\small{$0.8$}
                & $93.7\pm$\small{$1.7$}\\
                
                Satellite 
                & $65.6\pm$\small{$3.3$}
                & $74.6\pm$\small{$0.7$}
                & $\mathbf{77.8}\pm$\small{$0.4$}\\
                
                Pendigits 
                & $35.4\pm$\small{$10.9$}
                & $\mathbf{92.5}\pm$\small{$1.3$}
                & $\mathbf{93.1}\pm$\small{$1.2$}\\
                
                Annthyr. 
                & $29.9\pm$\small{$1.5$}
                & $\mathbf{57.7}\pm$\small{$0.6$} 
                & $19.6\pm$\small{$1.2$} \\
                
                Mnist 
                & $56.7\pm$\small{$5.7$}
                & $\mathbf{71.8}\pm$\small{$0.3$}
                & $69.7\pm$\small{$2.0$}\\
                
                Mammo. 
                & $17.4\pm$\small{$2.2$}
                & $\mathbf{43.6}\pm$\small{$0.5$}
                & $38.7\pm$\small{$1.7$} \\
                
                Shuttle 
                & $85.3\pm$\small{$9.8$}
                & $\mathbf{98.2}\pm$\small{$0.3$}
                & $95.2\pm$\small{$0.5$}\\
                
                Mulcross 
                & $\mathbf{100}\pm$\small{$0.0$}
                & $\mathbf{100}\pm$\small{$0.0$}
                & $\mathbf{100}\pm$\small{$0.0$} \\
                
                Forest 
                & N/A
                & $\mathbf{58.0}\pm$\small{$10$} 
                & $7.6\pm$\small{$1.2$}\\
                
                Campaign
                & $47.0\pm$\small{$1.9$}
                & $\mathbf{49.8}\pm$\small{$0.3$}
                & $42.0\pm$\small{$0.3$}\\
                
                Fraud 
                & $54.3\pm$\small{$5.2$}
                & $\mathbf{58.1}\pm$\small{$3.2$}
                & $41.8\pm$\small{$1.1$}\\
                
                Backdoor
                & $\mathbf{85.8}\pm$\small{$0.6$}
                & $84.1\pm$\small{$0.1$} 
                & $10.2\pm$\small{$0.6$}\\
                
                \midrule 
                
                mean 
                & $57.4$
                & $\mathbf{68.8}$
                & $57.5$\\
                
                \mbox{mean std} 
                & $4.3$
                & $\mathbf{2.0}$ 
                & $3.1$\\

                \mbox{mean rk} 
                & $8.0$ 
                & $\mathbf{3.4}$
                & $6.3$   \\
                \bottomrule
            \end{tabular}
        \end{center}
    \end{table*}
    \clearpage
    \begin{table*}[t!]
        \begin{center}
            \caption{Necessary dependencies to take into account for AD by mask-reconstruction. Given the obtained hierarchy between NPT-AD, Mask-KNN and the transformer on each dataset, we display the necessary dependencies to identify efficiently anomalies.}
            \label{tab:feature_or_sample}
            \begin{tabular}{lcc}
       \toprule
       Dependencies 
       & feature-feature
       & sample-sample 
       \\
       \midrule
       
        Wine 
        & $\checkmark$
        & $\checkmark$
        \\
        Lympho 
        & $\checkmark$
        & $\times$
        \\
        Glass 
        & $\times$
        & $\checkmark$
        \\
        
        Vertebral 
        & $\times$
        & $\checkmark$
        \\
        
        Wbc 
        & $\checkmark$
        & $\checkmark$
        \\
        
        Ecoli 
        & $\checkmark$
        & $\times$
        \\
        
        Ionosph. 
        & $\checkmark$
        & $\checkmark$
        \\
        
        Arrhyth. 
        & $\checkmark$
        & $\checkmark$
        \\
        
        Breastw 
        & $\checkmark$
        & $\checkmark$
        \\
        
        Pima 
        & $\checkmark$
        & $\checkmark$
        \\
        
        Vowels 
        & $\times$
        & $\checkmark$
        \\
        
        Letter 
        & $\checkmark$
        & $\checkmark$
        \\
        
        Cardio 
        & $\checkmark$
        & $\checkmark$
        \\
        
        Seismic 
        & $\times$
        & $\checkmark$
        \\
        
        Musk 
        & $\checkmark$
        & $\checkmark$
        \\
        
        Speech 
        & $\checkmark$
        & $\checkmark$
        \\
        
        Thyroid 
        & $\checkmark$
        & $\checkmark$
        \\
        
        Abalone 
        & $\checkmark$
        & $\checkmark$
        \\
        
        Optdigits 
        & $\times$
        & $\checkmark$
        \\
        
        Satimage
        & $\checkmark$
        & $\checkmark$
        \\
        
        Satellite 
        & $\times$
        & $\checkmark$
        \\
        
        Pendigits 
        & $\times$
        & $\checkmark$
        \\
        
        Annthyr. 
        & $\checkmark$
        & $\checkmark$
        \\
        
        Mnist 
        & $\times$
        & $\checkmark$
        \\
        
        Mammo. 
        & $\times$
        & $\checkmark$
        \\
        
        Shuttle 
        & $\checkmark$
        & $\checkmark$
        \\
        
        Mulcross 
        & $\checkmark$
        & $\checkmark$
        \\
        
        Forest 
        & $\checkmark$
        & $\times$
        \\

        Campaign
        & $\checkmark$
        & $\checkmark$
        \\
        
        Fraud 
        & $\checkmark$
        & $\times$
        \\
        
        Backdoor
        & $\checkmark$
        & $\times$
        \\
       \bottomrule
    \end{tabular}
        \end{center}
    \end{table*}
    \begin{table*}[h!]
    \begin{center}
    \caption{Mean rank (F1-score) for the experiments conducted, without Mask-KNN and with Mask-KNN}
        \label{tab:mean_rk_mask_knn}
        \begin{tabular}{lccc}
            \toprule
             Method & mean rank & mean rank (w/ Mask-KNN) & diff.  \\
             \midrule
             DROCC (abalone) & $11.5$ & $12.4$ & $+0.9$\\
             GOAD (thyroid) & $8.4$ & $9.1$ & $+0.7$\\
             NeuTraL-AD (arrhythmia) & $9.6$ & $10.3$ & $+0.7$\\
             Internal Cont. & $3.7$ & $4.0$ & $\mathbf{+0.3}$\\
             COPOD & $10.5$ & $11.0$ & $+0.5$\\
             IForest & $7.6$ & $8.1$ & $+0.5$\\
             KNN & $5.2$ & $5.6$ & $+0.4$\\
             PIDForest & $11.5$ & $12.2$ & $+0.7$ \\
             RRCF & $12.6$ & $13.4$ & $+0.8$ \\
             Transformer & $7.5$ & $8.0$ & $+0.5$ \\
             Mask-KNN & N/A & $6.6$ & N/A \\
             NPT-AD & $\mathbf{3.1}$ & $\mathbf{3.4}$ & $\mathbf{+0.3}$\\
             \bottomrule
        \end{tabular}
    \end{center}
    \end{table*}
    \clearpage
    
    \subsection{Computational time}
    \label{appendix:comput_time}
        \begin{table}[h!]
        \caption{Runtime in seconds of NPT-AD for the training and inference phase. The training runtime corresponds to the average training time of the model over the 20 runs with the parameters displayed in table \ref{tab:exp}. The inference runtime corresponds to the average runtime over the 20 runs to compute NPT-AD as shown in equation \ref{eq:npt-ad}.}
        \label{tab:runtime}
        \begin{center}
        \begin{tabular}{lcc}
           \toprule
           Dataset & train & inference \\
           \midrule
            Wine
            & $63$
            & $68$
            \\
            
            Lympho
            & $10$
            & $283$
            \\
            
            Glass
            & $76$
            & $6$
            \\
            
            Vertebral
            & $128$
            & $2$
            \\
            
            Wbc
            & $10$
            & $479$
            \\ 
            
            Ecoli
            & $11$
            & $23$
            \\
            
            Ionosph.
            & $12$
            & $76$
            \\
            
            Arrhyth.
            & $100$
            & $223$
            \\
            
            Breastw
            & $7$
            & $6$
            \\
            
            Pima
            & $37$
            & $18$
            \\
            
            Vowels
            & $62$
            & $63$
            \\
            
            Letter
            & $105$
            & $15$
            \\
            
            Cardio
            & $10$
            & $97$
            \\
            
            Seismic
            & $9$
            & $189$
            \\
            
            Musk
            & $56$
            & $168$
            \\
            
            Speech
            & $62$
            & $64$
            \\
            
            Thyroid
            & $253$
            & $2$
            \\
            
            Abalone
            & $55$
            & $50$
            \\
            
            Optdigits
            & $127$
            & $152$
            \\
            
            Satimage2
            & $13$
            & $17$
            \\
            
            Satellite
            & $13$ 
            & $23$
            \\
            
            Pendigits
            & $78$
            & $47$
            \\
            
            Annthyr.
            & $22$
            & $5$
            \\
            
            Mnist
            & $478$
            & $153$
            \\
            
            Mammo.
            & $16$
            & $24$
            \\
            
            Shuttle
            & $16$
            & $115$
            \\
            
            Mullcross
            & $43$
            & $44$
            \\
            
            Forest
            & $73$
            & $409$
            \\

            Campaign
            & $52$
            & $251$ \\

            Fraud
            & $141$
            & $362$ \\

            Backdoor
            & $18396$
            & $1992$
            \\
           \bottomrule
        \end{tabular}
        \end{center}
        \end{table}

\end{appendices}

\end{document}